\documentclass{article} 
\usepackage{iclr2027_conference,times}

\iclrfinalcopy

\usepackage{amsmath,amsfonts,bm}

\def\eqref#1{equation~\ref{#1}}

\def\1{\bm{1}}

\DeclareMathAlphabet{\mathsfit}{\encodingdefault}{\sfdefault}{m}{sl}
\SetMathAlphabet{\mathsfit}{bold}{\encodingdefault}{\sfdefault}{bx}{n}

\usepackage{hyperref}
\usepackage{url}
\usepackage{microtype}
\usepackage{graphicx}
\usepackage{subcaption}
\usepackage{booktabs}
\usepackage{multirow}
\usepackage{wrapfig}
\usepackage{amsmath,amssymb,amsfonts,amsthm,mathtools,bm}
\usepackage{siunitx}
\usepackage{nicefrac}
\usepackage{textcomp}
\usepackage{url}
\usepackage{xcolor}
\usepackage{hyperref}
\usepackage{enumitem}
\usepackage{wrapfig}
\usepackage{booktabs}
\usepackage{multirow}
\usepackage[table]{xcolor}
\usepackage{graphicx}
\usepackage{caption}
\usepackage[most]{tcolorbox}
\usepackage{algorithm}
\usepackage{algpseudocode}
\usepackage{amsmath}

\usepackage{microtype}
\usepackage{hyperref}
\usepackage{url}
\usepackage{graphicx}
\usepackage{wrapfig}
\usepackage{amssymb}
\usepackage{amsthm}
\usepackage{amsmath}
\usepackage{booktabs}
\usepackage{makecell}
\usepackage{multirow}
\usepackage{caption}
\usepackage[capitalize,noabbrev]{cleveref}
\usepackage{subcaption}
\usepackage{enumitem}
\usepackage{adjustbox}
\usepackage{pgfplots}
\usepackage{titletoc}
\usepackage[most]{tcolorbox} %
\tcbuselibrary{listings,skins,breakable}
\usepackage{listings}
\usepackage[T1]{fontenc}        %
\usepackage{fontawesome5}    %
\usepackage{multirow}
\usepackage{xcolor,colortbl}
\usepackage{arydshln}
\usepackage{svg}
\usepackage[frozencache]{minted}
\usepackage{xcolor}

\usepackage{booktabs}
\usepackage{multirow}
\usepackage{graphicx}

\usepackage{lipsum}

\usepackage{tikz}
\usetikzlibrary{arrows.meta,positioning,fit,backgrounds,calc}

\definecolor{sfgray}  {RGB}{238,240,243}
\definecolor{sfblue}  {RGB}{219,233,246}
\definecolor{sfgreen} {RGB}{225,240,222}
\definecolor{sfamber} {RGB}{253,238,214}
\definecolor{sfviolet}{RGB}{232,227,244}
\definecolor{sfyellow}{RGB}{255,248,219}
\definecolor{sfrose}  {RGB}{250,228,228}

\newcommand{\ourform}{Function-Space Policy Optimization}
\newcommand{\our}{FSPO}

\title{Function-Space Temporal Self-Distillation for Policy Optimization}

\author{%
\textbf{Aryan Das}$^{*1}$  \quad
\textbf{Arnab Bhattacharya}$^{1}$ \quad
 \textbf{Vinay Kumar Verma}$^{*1}$ 
\\[7pt]
\small
$^{1}$ Indian Institute of Technology Kanpur, India\\
}

\begin{document}

\maketitle

\begin{abstract}

Reinforcement Learning with Verifiable Rewards (RLVR) improves reasoning in language models, but scalar outcomes provide coarse supervision over response trajectories. Self-distillation offers denser token-level guidance by comparing an unassisted policy with a feedback-informed self-teacher, yet the teacher evolves with the student, creating unstable targets and motivating additional regularization. We introduce {\ourform{} (\our{})}, a single-policy framework that combines current privileged adaptation with function-space temporal consolidation. \our{} stores sparse historical next-token distributions as fixed temporal targets, providing temporal separation without maintaining a persistent teacher model. It further introduces privileged disagreement correction to explicitly suppress locally overvalued actions, reliability-adaptive grounding that strengthens reference supervision when self-generated evidence repeatedly fails, and quality- and coverage-aware temporal memory that retains verified behaviors and improves privileged demonstrations. Across reasoning and tool-use benchmarks, \our{} achieves the highest average accuracy under a 5-hour training budget, improving over the strongest baseline by approximately 4\% points on average while reducing peak memory from about 70\,GB to 56\,GB on Olmo3-7B-Instruct.
\begingroup
\renewcommand\thefootnote{}\footnotetext{\textsuperscript{*} Equal Contributions}
\addtocounter{footnote}{-1}
\endgroup

\end{abstract}    
\section{Introduction}
\label{sec:intro}

\begin{wrapfigure}{r}{0.45\textwidth}
  \vspace{-5mm}
  \centering
  \includegraphics[width=\linewidth]{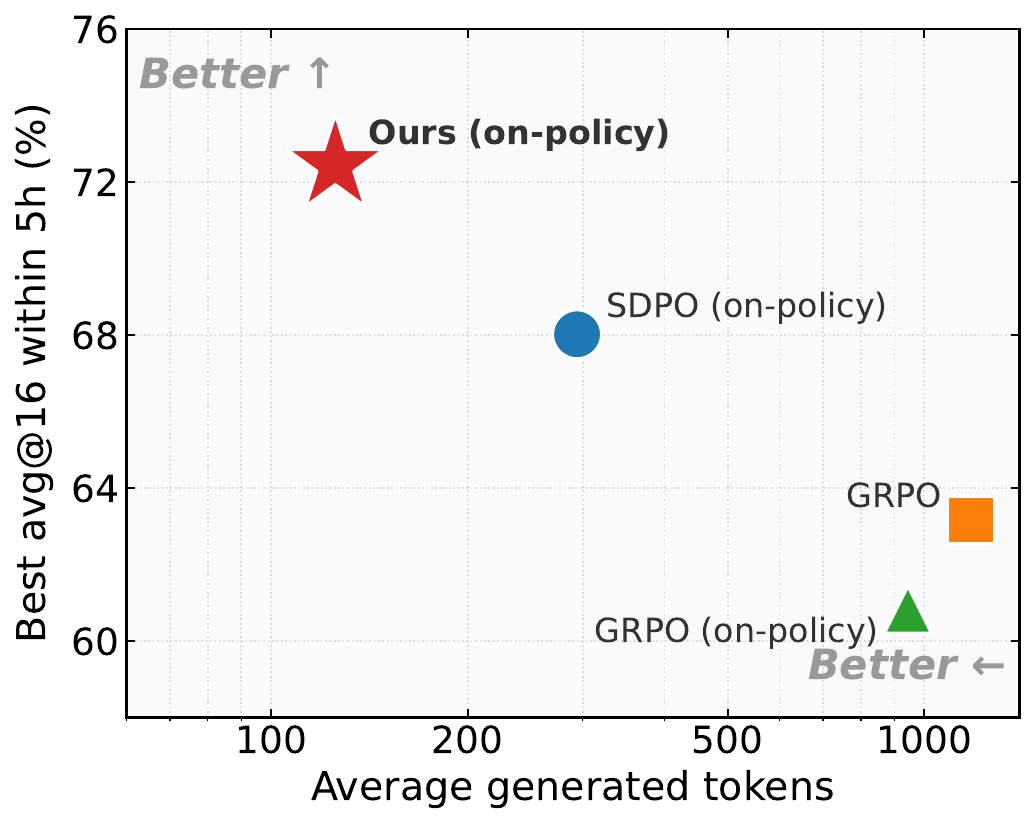}
  \vspace{-7mm}
  \caption{Average Generated Tokens vs.\ avg@16 Accuracy (Olmo3-7B-Instruct, 5h) across five tasks.
  Ours reaches the highest accuracy with the fewest tokens.}
  \label{fig:tokens_vs_acc}
  \vspace{-3mm}
\end{wrapfigure}

RLVR has become a paradigm for post-training Large Language Models (LLMs), driving substantial improvements in reasoning, problem solving, code generation, and tool use~\cite{shao2024deepseekmath,team2025kimi,lambert2024tulu}. Most RLVR methods still reduce an entire response to a single scalar outcome, despite decisions of varying quality along the trajectory.
This creates a difficult credit-assignment problem, particularly when rewards are sparse or when all sampled responses for a problem receive the same outcome~\cite{kazemnejad2024vineppo,sdpo}. Recent self-distillation methods offer a compelling alternative by extracting richer supervision from the model's own behavior rather than relying solely on scalar advantages or a separate expert teacher~\cite{agarwal2024policy,shenfeld2026self,zhao2026self}. 
In particular, Self-Distillation Policy Optimization (SDPO)~\cite{sdpo} shows that a single policy can serve as both student and self-teacher, first generating a response independently and then re-evaluating it using additional feedback.
The difference between these views provides token-level supervision for finer credit assignment than outcome rewards.

SDPO has several limitations that arise from its reliance on a continuously evolving self-teacher. Because the student and self-teacher are instantiated from the same policy, the teacher distribution changes whenever the policy parameters are updated. Although stop-gradient prevents gradient flow through the teacher within an optimization step, it does not provide temporal separation across training iterations, creating a moving-target problem that can weaken the stability of the distillation signal. SDPO therefore relies on stabilization mechanisms such as EMA or trust-region regularization; the EMA variant requires an additional persistent parameter copy, introducing memory overhead that grows with model size. SDPO also treats student–teacher disagreement primarily through distribution matching and does not explicitly distinguish ordinary mismatch from cases where the student strongly overvalues a candidate rejected by the privileged view. Such severe local disagreements may require more targeted directional correction. In addition, SDPO lacks a mechanism for adapting supervision according to the reliability of the policy’s behavior on individual problems. Examples that the policy repeatedly fails to solve are therefore not explicitly differentiated, in supervision strength, from those for which reliable self-generated guidance is consistently available. Finally, SDPO primarily relies on the current feedback-conditioned policy and does not maintain an explicit temporal memory of useful privileged predictions discovered earlier in training. As a result, informative past behaviors are not systematically preserved and replayed, nor are historical experiences selected to balance solution quality with structural diversity. These limitations restrict stable temporal supervision, targeted local correction, reliability-aware learning, and preservation of complementary behaviors across training.

\vspace{-1mm}
\begin{tcolorbox}[
enhanced,
colback=yellow!8!white,
colframe=black,
boxrule=0.7pt,
arc=2mm,
width=\linewidth,
top=1mm,
bottom=1mm,
left=1mm,
right=1mm
]
\centering
\itshape\bfseries
Can we achieve temporally stable and reliability-aware self-distillation without maintaining a separate teacher model, while preserving useful knowledge from earlier stages of training?
\end{tcolorbox}
\vspace{-1mm}

\our{} addresses these limitations through a unified self-distillation framework that improves temporal stability, strengthens local correction, adapts supervision to policy reliability, and preserves useful historical knowledge. To mitigate the moving-target problem, informative privileged predictions from earlier optimization steps are materialized as sparse next-token distributions and retained as fixed temporal targets. Because these historical function values remain unchanged while the current policy evolves, they provide temporal separation across training iterations without requiring a persistent EMA teacher or an additional resident parameter copy. \our{} also goes beyond distribution-level matching by identifying candidates that the unassisted policy strongly prefers but the privileged view rejects, and explicitly suppressing these locally overvalued decisions. This distinguishes severe privileged disagreement from ordinary distribution mismatch and provides a more targeted directional learning signal. Supervision is further adapted according to the reliability of the policy on each problem over time. A smoothed estimate of rollout performance distinguishes problems on which the policy repeatedly fails from those for which it consistently produces reliable solutions. When the policy repeatedly fails and its own evidence becomes unreliable, the reference loss is strengthened; as the policy improves and begins to solve the problem consistently, this supervision is progressively reduced. \our{} additionally maintains a temporal memory of useful privileged behaviors discovered earlier in training. This memory provides fixed historical targets for consolidation and stronger demonstrations when the current rollout group lacks a verified successful response. By retaining trajectories according to both solution quality and structural complementarity, \our{} preserves diverse, nonredundant solution strategies and enables the policy to benefit from complementary historical knowledge throughout optimization.
As shown in Figure~\ref{fig:tokens_vs_acc}, these design choices translate into the highest average accuracy across the five tasks while requiring the fewest generated tokens.
Our contributions are summarized as follows:
\vspace{-3mm}
\begin{itemize}[leftmargin=10pt]
\item Preserves temporal knowledge through sparse historical next-token distributions while combining current privileged adaptation with historical function-space consolidation, enabling temporally separated supervision with a single resident policy.
\item Introduces targeted privileged disagreement correction that suppresses candidates strongly overvalued by the unassisted policy but rejected by the privileged view, providing stronger directional correction beyond distribution matching.
\item Adapts reference supervision to the reliability of self-generated guidance, strengthening it when the policy repeatedly fails and reducing it as reliable solutions emerge. Quality- and coverage-aware temporal memory further preserves diverse, nonredundant behaviors.
\item Achieves approximately 4\% points higher average accuracy than the strongest baseline across reasoning and tool-use benchmarks, while reducing peak memory from about 70\,GB to 56\,GB for Olmo3-7B-Instruct by eliminating the persistent EMA teacher.
\end{itemize}
\vspace{-4mm}

\section{\ourform}
\label{sec:ftsd}

\begin{figure*}[t]
    \vspace{-2mm}
    \centering
    \begin{minipage}[c]{0.64\linewidth}
        \centering
        \includegraphics[width=\linewidth]{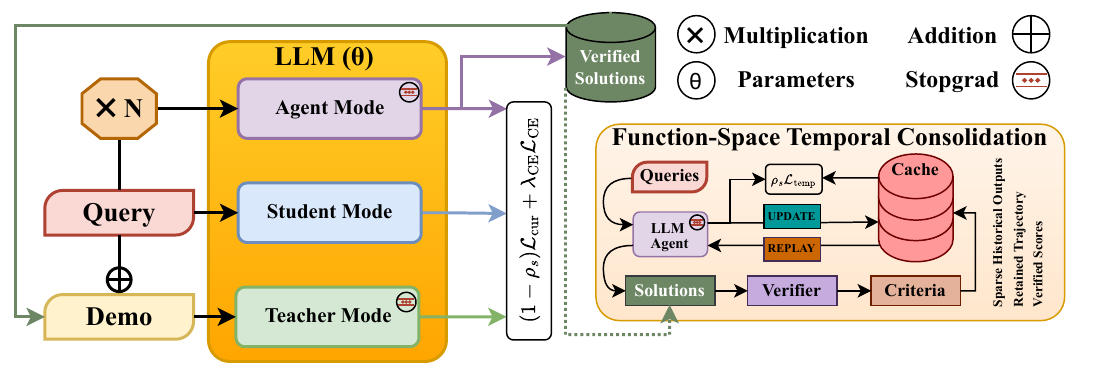}
        \caption{
        \textit{\our{} Overview.}
        For each query, the policy samples rollouts in Agent mode; a successful verified rollout is used as a demo. The Student mode predicts from the query alone, while the Teacher mode conditions on the query with a demo, producing token-level supervision. Verified solutions are then stored as sparse function-space targets in a cache and later replayed for temporal consolidation, preserving useful past behavior without maintaining a separate parameter-space teacher.
        }
        \label{fig:method}
    \end{minipage}
    \hfill
    \begin{minipage}[c]{0.34\linewidth}
        \vspace{-4mm}
        \algrenewcommand\algorithmicindent{0.8em}
        \begin{algorithm}[H]
        \scriptsize
        \caption{\our{} Training}
        \label{alg:ftsd}
        \begin{algorithmic}[1]
        \Require $\pi_{\theta}$, $\mathcal{D}$, $\hat r_p\gets 0$, memory $\mathcal{M}\gets$ reference seeds
        \For{$s=1,\ldots,T$}
            \State Draw $G$ rollouts $y_{p,j}$, score $r_{p,j}$
            \State Demo $d_{p,j}$: sibling $\rightarrow\tilde y_p\rightarrow g_p$
            \State Student $p_{p,j,t}$, teacher $q_{p,j,t}$
            \State Keep top-$K_{\mathrm{cur}}$ support $S_{p,j,t}$
            \State Overvalued set $\mathcal{N}_{p,j,t}$ via margin $\gamma$
            \State $\mathcal{L}_{\mathrm{cur}}\gets\mathcal{L}_{\mathrm{distill}}+\lambda_{\mathrm{neg}}\mathcal{L}_{\mathrm{neg}}$
            \State Update $\hat r_p^{(s)}$, $w_p^{(s)}$; get $\mathcal{L}_{\mathrm{CE}}$
            \State Replay $\mathcal{R}_s\subset\mathcal{M}$ by group
            \State $\mathcal{L}_{\mathrm{temp}}$ vs.\ fixed targets $\tilde q_{p,t}$
            \State Set $\rho_s$ by cyclic schedule
            \State Update $\theta_s$ by minimizing $\mathcal{L}(\theta_s)$
            \State Retain verified rollouts (quality/coverage)
            \State Store $(\tilde S_{p,t},\tilde q_{p,t})$ in $\mathcal{M}$
            \If{$s \bmod F = 0$}
                \State Refresh $\zeta$ of entries aged $>H$
            \EndIf
        \EndFor
        \end{algorithmic}
        \end{algorithm}
    \end{minipage}
    \vspace{-5mm}
\end{figure*}

Self-distillation provides dense supervision by comparing an unassisted policy with a privileged self-teacher conditioned on stronger evidence. However, because both views evolve with the same policy, the teacher remains a moving target across optimization steps. FTSD addresses this by preserving informative historical policy outputs directly in function space rather than maintaining a persistent parameter-space teacher. Let $\pi_{\theta_s}$ denote the policy at step $s$, with
\begin{equation}
    F_{\theta_s}(c)=\pi_{\theta_s}(\cdot\mid x,y_{<t})
\end{equation}
for prefix $c=(x,y_{<t})$. Selected privileged predictions are materialized as sparse next-token distributions and retained as temporal targets, providing temporal separation as the policy evolves. 
FTSD further combines current privileged adaptation with targeted disagreement correction for strongly overvalued actions, quality- and coverage-aware temporal consolidation of complementary historical behaviors, and reliability-adaptive reference grounding. Reference supervision is strengthened when self-generated evidence fails and progressively reduced as reliable behavior emerges, yielding stable, selective, and adaptive self-distillation within a single-policy framework.
Figure~\ref{fig:method} illustrates the overall framework, while Algorithm~\ref{alg:ftsd} summarizes the complete training procedure.

\subsection{Privileged Adaptation and Disagreement Correction}

We first construct dense supervision by comparing the current policy under an unassisted view and a privileged view with access to stronger evidence. Let $\mathcal{D}=\{(x_p,g_p)\}_{p=1}^{N}$ denote the training set, where $g_p$ is the verified reference solution. For each problem, the policy samples $G$ responses,
\begin{equation}
    y_{p,j}\sim\pi_{\theta_s}(\cdot\mid x_p),
\qquad
r_{p,j}=R(x_p,y_{p,j},g_p)\in[0,1].
\end{equation}

An unsuccessful rollout reveals where the policy fails, but not how its local token decisions should change. We therefore construct a privileged demonstration using the strongest reliable evidence available:
\begin{equation}
    d_{p,j}=
\begin{cases}
y_{p,k}, & \exists\,k\neq j:\; r_{p,k}\ge\tau_{\mathrm{succ}},\\
\tilde y_p, & \text{if a verified temporal trajectory is available},\\
g_p, & \text{otherwise}.
\end{cases}
\end{equation}

This hierarchy, with success threshold $\tau_{\mathrm{succ}}>0$, prioritizes a verified sibling response, followed by previously retained policy behavior, while using the reference solution as a reliable fallback. At each response position $t$, the unassisted student and privileged self-teacher evaluate the same response prefix,
\begin{equation}
    p_{p,j,t}
=
\pi_{\theta_s}(\cdot\mid x_p,y_{p,j,<t}),
\qquad
q_{p,j,t}
=
\operatorname{stopgrad}
\left[
\pi_{\theta_s}(\cdot\mid x_p\oplus d_{p,j},y_{p,j,<t})
\right].
\end{equation}

Since both views share the same policy and response history, their disagreement reflects how privileged evidence changes the next-token prediction at the same response state.

To avoid matching the full vocabulary distribution, we retain the $K_{\mathrm{cur}}$ highest-probability student tokens and aggregate the remaining probability mass into a tail category:
\begin{equation}
    \Phi_S(a)
=
\left(
\{a(v)\}_{v\in S},
1-\sum_{v\in S}a(v)
\right).
\end{equation}
where a is a probability distribution over the vocabulary,  a(v) is the probability that distribution assigns to token v and S is a fixed subset of the vocabulary. Specifically it's the top-K highest-probability tokens under the student's distribution. $\Phi_S(a)$ reduces the full distribution a down to: its values on that subset $\{a(v)\}_{v\in S}$, plus one extra "tail" bucket holding all the leftover mass $1-\sum_{v\in S}a(v)$.

Let $S_{p,j,t}$ denote the student support and $M_{p,j,t}$ indicate valid response positions. We transfer privileged information using generalized Jensen-Shannon divergence (JSD),
\begin{equation}
    \mathcal{L}_{\mathrm{distill}}
=
\frac{1}{Z_{\mathrm{cur}}}
\sum_{p,j,t}
M_{p,j,t} \times
JSD
\left(
\Phi_{S_{p,j,t}}(p_{p,j,t}),
\Phi_{S_{p,j,t}}(q_{p,j,t})
\right),
\qquad
Z_{\mathrm{cur}}
=
\sum_{p,j,t}M_{p,j,t}.
\end{equation}

This provides a distinct privileged target at every response position, yielding dense local supervision rather than a single outcome-level signal.

Some disagreements, however, provide a stronger directional signal. When the student assigns substantially more probability to a token than the privileged view, that candidate is locally overvalued. We identify such tokens as
\begin{equation}
    \mathcal{N}_{p,j,t}
=
\left\{
v\in S_{p,j,t}:
\log p_{p,j,t}(v)-\log q_{p,j,t}(v)>\gamma
\right\},
\end{equation}

and penalize them using
\begin{equation}
    \ell_{\mathrm{neg}}(p,j,t)
=
\frac{1}{\max(1,|\mathcal{N}_{p,j,t}|)}
\sum_{v\in\mathcal{N}_{p,j,t}}
\left[
\log p_{p,j,t}(v)
-
\operatorname{stopgrad}(\log q_{p,j,t}(v))
-
\gamma
\right]_{+},
\end{equation}

\begin{equation}
    \mathcal{L}_{\mathrm{neg}}
=
\frac{1}{Z_{\mathrm{cur}}}
\sum_{p,j,t}
M_{p,j,t}\,
\ell_{\mathrm{neg}}(p,j,t).
\end{equation}

The privileged adaptation objective is therefore
\begin{equation}
    \mathcal{L}_{\mathrm{cur}}
=
\mathcal{L}_{\mathrm{distill}}
+
\lambda_{\mathrm{neg}}\mathcal{L}_{\mathrm{neg}}.
\end{equation}

This provides dense token-level supervision while explicitly correcting local decisions that are strongly contradicted by privileged evidence.

\vspace{-2mm}
\subsection{Function-Space Temporal Consolidation}

Privileged adaptation provides supervision at the current optimization step, but its target evolves together with the policy. Thus, stop-gradient separates the student and self-teacher within an update, but does not provide temporal separation across updates. We address this by preserving selected privileged predictions from earlier stages as fixed function-space targets rather than maintaining a persistent historical teacher.

For each problem $p$, the temporal memory stores a retained trajectory $\tilde y_p$, its verifier score $\tilde r_p$, structural signature $\sigma_p$, and sparse privileged predictions $\{\tilde S_{p,t},\tilde q_{p,t}\}_t$. To improve the usefulness of this memory, trajectories are retained based on both verifier quality and structural complementarity rather than reward alone. This discourages repeatedly storing similar solution patterns and promotes broader coverage of distinct behaviors. A candidate replaces the stored trajectory when its retention score, which combines verifier quality with the coverage of its structural signature, exceeds that of the stored entry. During consolidation, memories are also sampled evenly across structural groups, preventing dominant signatures from overwhelming replay.

Let $\theta_p^{\mathrm{src}}$ denote the policy state from which a temporal target is materialized. We use $\tilde d_p=\tilde y_p$ when $\tilde r_p\ge\tau_{\mathrm{succ}}$ and $\tilde d_p=g_p$ otherwise. The source policy produces
\vspace{-2mm}
\begin{equation}
    \bar q_{p,t}
=
\operatorname{stopgrad}
\left[
\pi_{\theta_p^{\mathrm{src}}}
(\cdot\mid x_p\oplus\tilde d_p,\tilde y_{p,<t})
\right].
\end{equation}
Rather than retaining the source parameters, we store only its dominant predictions, \vspace{-2mm}
\begin{equation}
    \tilde S_{p,t}
=
\operatorname{TopK}(\bar q_{p,t},K_{\mathrm{temp}}),
\qquad
\tilde q_{p,t}
=
\Phi_{\tilde S_{p,t}}(\bar q_{p,t}),
\end{equation}

which remain fixed while the current policy evolves.

At optimization step $s$, let $\mathcal{R}_s$ denote the temporal memories selected for consolidation. The current policy evaluates each retained trajectory under the original problem context and is matched against its fixed historical privileged target:
\begin{equation}
\small
    \mathcal{L}_{\mathrm{temp}}
=
\frac{1}{Z_{\mathrm{temp}}}
\sum_{p\in\mathcal{R}_s}\sum_t
\tilde M_{p,t}
D_{\alpha}
\left(
\Phi_{\tilde S_{p,t}}
\left[
\pi_{\theta_s}(\cdot\mid x_p,\tilde y_{p,<t})
\right],
\tilde q_{p,t}
\right),
\qquad
Z_{\mathrm{temp}}
=
\sum_{p\in\mathcal{R}_s,t}\tilde M_{p,t}.
\end{equation}

Unlike the current privileged target, $\tilde q_{p,t}$ remains fixed while the policy evolves. Privileged adaptation therefore captures what should be learned from current evidence, whereas temporal consolidation preserves behavior that was informative at earlier stages of training.

To avoid stale targets without collapsing temporal separation, we use delayed and staggered refresh. A stored target becomes eligible for refresh after $H$ optimization steps, and a fraction $\zeta$ of eligible entries is updated every $F$ steps. This maintains temporal targets of different ages while allowing the memory to gradually track policy improvement.


\subsection{Reliability-Adaptive Reference Grounding}

Privileged adaptation and temporal consolidation primarily rely on supervision generated by the policy itself. However, this signal can be unreliable for problems that the policy repeatedly fails to solve. We therefore use the reference solution as an adaptive grounding signal whose strength depends on the reliability of self-generated behavior.

For problem $p$, we measure the current rollout reliability and its smoothed estimate as
\begin{equation}
    r_p^{(s)}
=
\frac{1}{G}\sum_{j=1}^{G} r_{p,j},
\qquad
\hat r_p^{(s)}
=
(1-\nu)\hat r_p^{(s-1)}
+
\nu r_p^{(s)},
\end{equation}

where $\nu$ controls the influence of recent outcomes. Given a target reliability $\tau$, the reference weight is
\begin{equation}
    w_p^{(s)}
=
\operatorname{clip}
\left(
\frac{\tau-\hat r_p^{(s)}}{\tau},
0,1
\right).
\end{equation}

Thus, reference supervision is strongest when the policy repeatedly fails and its own evidence remains unreliable, and gradually decreases as the policy begins to solve the problem consistently.

Let $g_p=(g_{p,1},\ldots,g_{p,L_p})$ be the tokenized reference solution, $\mathcal{B}_s^{\mathrm{CE}}=\{p\in\mathcal{B}_s:w_p^{(s)}>0\}$ the grounded problems, and $A_{p,t}\in\{0,1\}$ a mask over valid reference tokens excluding prompt and padding positions. The grounding objective is
\begin{equation}
    \mathcal{L}_{\mathrm{CE}}
=
-\frac{1}{Z_{\mathrm{CE}}}
\sum_{p\in\mathcal{B}_s^{\mathrm{CE}}}\sum_t
w_p^{(s)}A_{p,t}
\log\pi_{\theta_s}(g_{p,t}\mid x_p,g_{p,<t}),
\qquad
Z_{\mathrm{CE}}
=
\sum_{p\in\mathcal{B}_s^{\mathrm{CE}},t}A_{p,t}.
\end{equation}

The reference therefore acts as a reliability-dependent grounding signal rather than uniform supervised training: it supports problems for which self-generated evidence is weak and progressively vanishes as reliable self-supervision emerges.


\subsection{Unified Adaptation and Consolidation Objective}

The complete objective combines current privileged adaptation, historical consolidation, and reliability-adaptive grounding:
\begin{equation}
    \mathcal{L}(\theta_s)
=
(1-\rho_s)\mathcal{L}_{\mathrm{cur}}
+
\rho_s\mathcal{L}_{\mathrm{temp}}
+
\lambda_{\mathrm{CE}}\mathcal{L}_{\mathrm{CE}},
\end{equation}


Here, $\rho_s\in[0,1]$ controls the balance between acquiring current privileged information and consolidating useful historical behavior, while $\lambda_{\mathrm{neg}}$ and $\lambda_{\mathrm{CE}}$ determine the strengths of disagreement correction and reference grounding.

Rather than fixing this balance throughout training, we periodically shift emphasis from adaptation toward consolidation:
\begin{equation}
    \rho_s
=
\rho_{\min}
+
(\rho_{\max}-\rho_{\min})
\frac{(s-1)\bmod C}{\max(1,C-1)},
\end{equation}

where $C$ is the cycle length. Since memory is pre-seeded with reference trajectories, temporal consolidation starts immediately. Each cycle initially favors current privileged supervision and progressively increases the weight of historical targets, balancing adaptation with retention.
\vspace{-3mm}
\section{Experiments}
\label{sec:experiments}
\vspace{-2mm}


\begin{table}[t]
\centering
\vspace{-2mm}
\caption{
Comparison of \our{} with GRPO and SDPO across benchmarks. We report the highest avg@16 achieved within 1-hour and 5-hour wall-clock training budgets for Qwen3-8B and Olmo3-7B-Instruct, with baseline results for GRPO, on-policy GRPO, and SDPO taken from SDPO~\cite{sdpo}. 
The final column reports the average performance across all five benchmarks at the 5-hour budget.
\our{} follows the SDPO training and evaluation setting on 4 NVIDIA GH200 GPUs.
}
\label{tab:table1}
\small
\vspace{-2mm}

\resizebox{\linewidth}{!}{%
\begin{tabular}{l *{10}{c} c}
\toprule
& \multicolumn{2}{c}{Chemistry}
& \multicolumn{2}{c}{Physics}
& \multicolumn{2}{c}{Biology}
& \multicolumn{2}{c}{Materials}
& \multicolumn{2}{c}{Tool use}
& \textbf{Avg@5h} \\
\cmidrule(lr){2-3}
\cmidrule(lr){4-5}
\cmidrule(lr){6-7}
\cmidrule(lr){8-9}
\cmidrule(lr){10-11}
\cmidrule(lr){12-12}
& 1h & 5h
& 1h & 5h
& 1h & 5h
& 1h & 5h
& 1h & 5h
& \\
\midrule

\textbf{Qwen3-8B}
& \multicolumn{2}{c}{41.2}
& \multicolumn{2}{c}{59.2}
& \multicolumn{2}{c}{30.8}
& \multicolumn{2}{c}{58.9}
& \multicolumn{2}{c}{57.5}
& 49.5 \\

\quad + GRPO
& 65.9 & 74.5
& 63.8 & 72.7
& 35.1 & 59.9
& 74.3 & 77.1
& 64.9 & 67.7
& 70.4 \\

\quad + GRPO {\small (on-policy)}
& 63.3 & 63.4
& 63.6 & 63.6
& 49.8 & 49.8
& 73.9 & 74.1
& 60.2 & 65.7
& 63.3 \\

\quad + SDPO {\small (on-policy)}
& 73.2 & 80.9
& 66.6 & 75.6
& 50.6 & 56.8
& 72.1 & 78.4
& \textbf{68.0} & \textbf{68.5}
& 72.0 \\

\quad + \textbf{\our{}} {\small (on-policy)}
& \textbf{81.8} & \textbf{83.4}
& \textbf{66.7} & \textbf{76.0}
& \textbf{71.2} & \textbf{71.2}
& \textbf{81.7} & \textbf{81.7}
& 66.6 & \textbf{68.5}
& \textbf{76.2} \\

\midrule

\textbf{Olmo3-7B-Instruct}
& \multicolumn{2}{c}{22.8}
& \multicolumn{2}{c}{37.7}
& \multicolumn{2}{c}{16.2}
& \multicolumn{2}{c}{36.7}
& \multicolumn{2}{c}{39.3}
& 30.5 \\

\quad + GRPO
& 39.7 & 56.7
& 55.3 & 63.3
& 35.6 & 55.8
& 70.9 & 75.0
& 56.4 & 65.0
& 63.2 \\

\quad + GRPO {\small (on-policy)}
& 51.4 & 57.5
& 62.7 & 62.7
& 49.8 & 49.8
& 73.3 & 73.5
& 56.8 & 60.6
& 60.8 \\

\quad + SDPO {\small (on-policy)}
& 68.0 & 80.0
& 59.9 & 66.1
& 48.0 & 52.8
& 73.7 & 79.1
& \textbf{60.8} & 62.1
& 68.0 \\

\quad + \textbf{\our{}} {\small (on-policy)}
& \textbf{81.0} & \textbf{83.3}
& \textbf{66.4} & \textbf{70.1}
& \textbf{62.3} & \textbf{62.3}
& \textbf{76.3} & \textbf{79.9}
& 57.6 & \textbf{66.6}
& \textbf{72.4} \\

\bottomrule
\end{tabular}%
}
\vspace{-2mm}
\end{table}

\begin{figure*}[t]
    \centering
    \includegraphics[width=\linewidth]{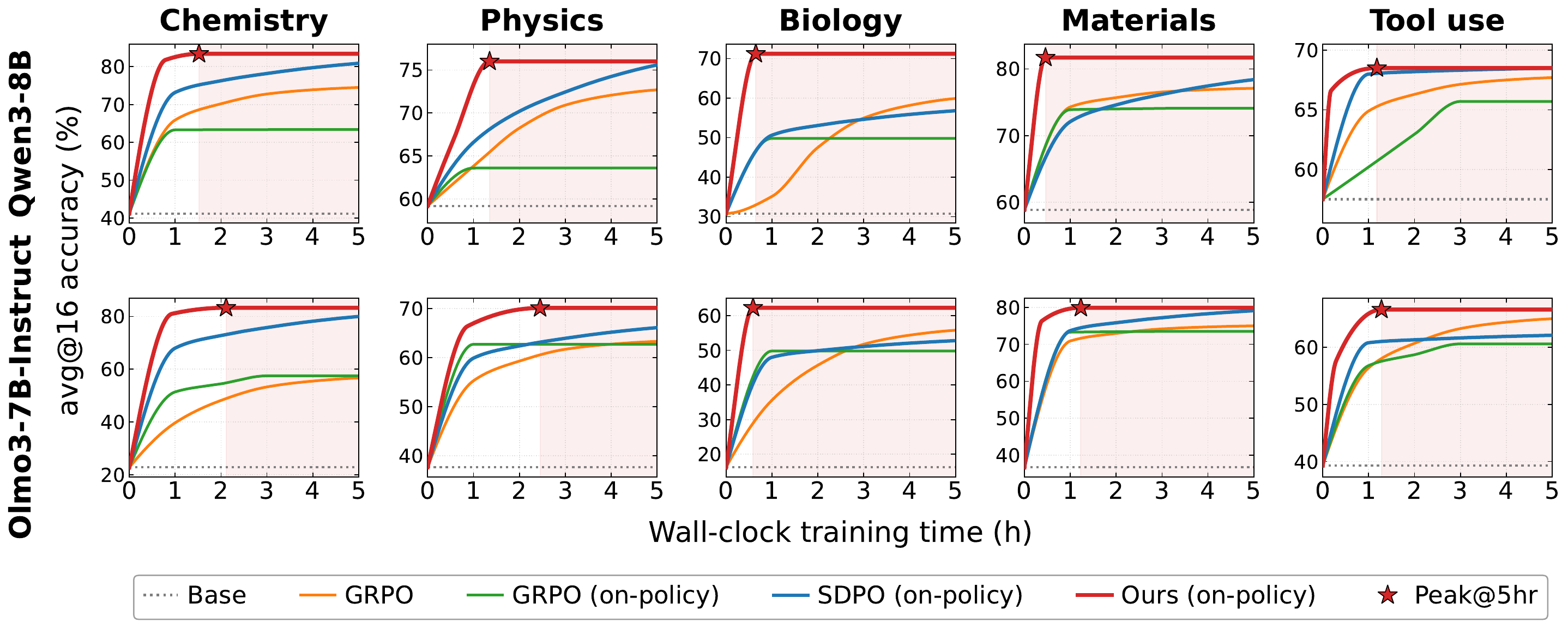}
    
    \vspace{-2mm}
    \caption{
    Convergence of best-within-window avg@16 accuracy over wall-clock training time for Qwen3-8B (top row) and Olmo3-7B-Instruct (bottom row) across five tasks.
    }
    \label{fig:convergence}
    \vspace{-5mm}
\end{figure*}

\begin{figure*}[t]
    \vspace{-2mm}
    \centering
    \includegraphics[width=\textwidth]{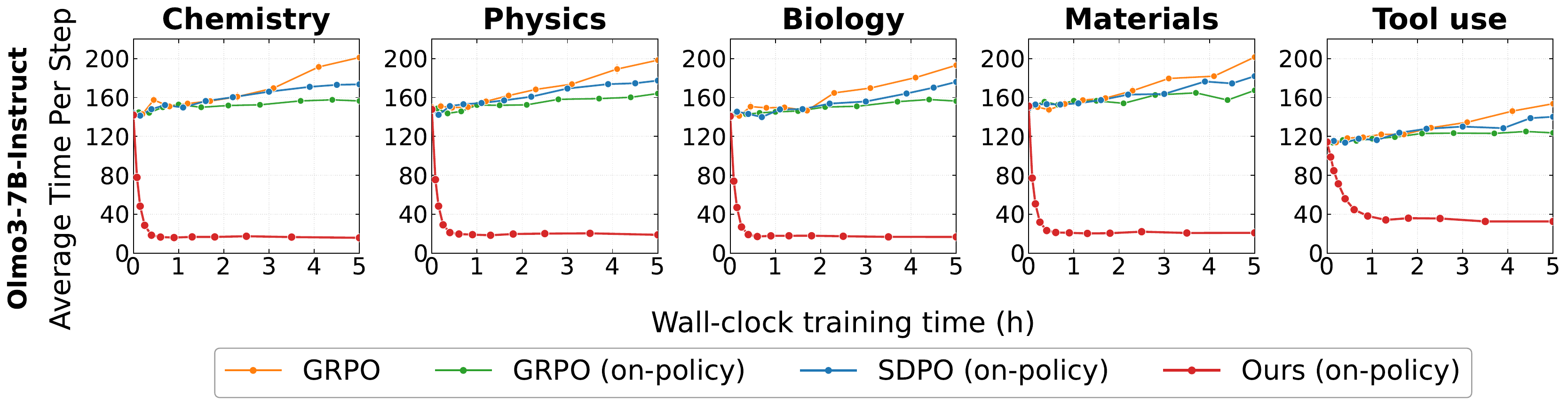}

    \vspace{-2mm}
    \caption{
    Average time per training step over wall-clock time for Olmo3-7B-Instruct across five tasks.
    }
    \label{fig:time_per_step}
    \vspace{-4mm}
\end{figure*}

\begin{figure*}[t]
\centering
\includegraphics[width=\textwidth]{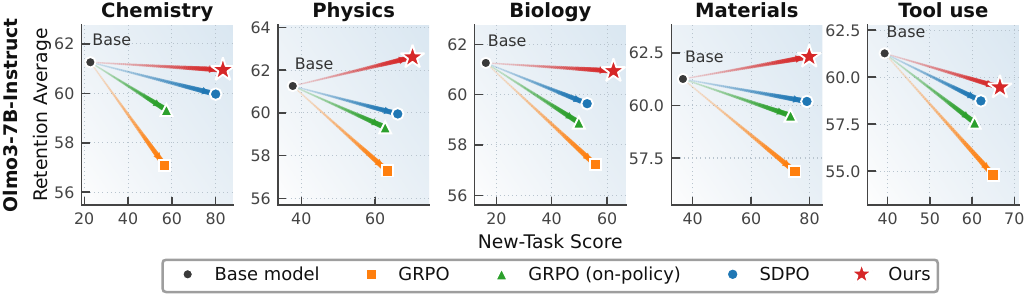}

\vspace{-2mm}
\caption{
Catastrophic-forgetting trade-off on Olmo3-7B-Instruct.
Each panel corresponds to training on one task, with the horizontal axis showing the best avg@16 achieved on that task within 5\,h and the vertical axis showing the mean score on the $5$ prior-capability benchmarks. 
}
\label{fig:forgetting}
\vspace{-5mm}
\end{figure*}

\begin{figure*}[t]
    \vspace{-2mm}
    \centering
    \includegraphics[width=0.9\textwidth, height=4.5cm]{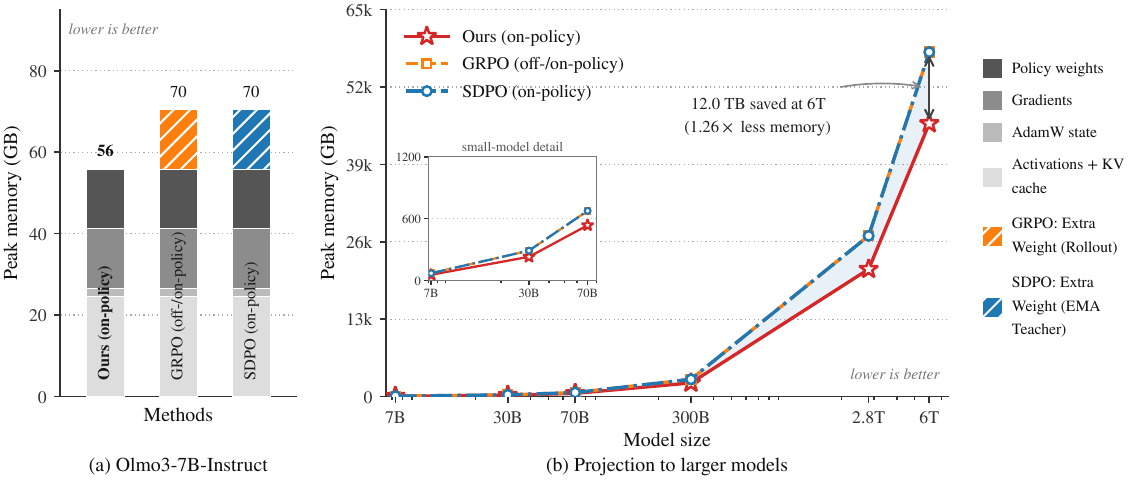}

    \vspace{-2mm}
    \caption{Comparison of Memory Footprint. (a) Breakdown for Olmo3-7B-Instruct: policy weights, gradients, AdamW state, and activations plus KV cache are common to all methods. GRPO adds a synced colocated rollout copy, and SDPO adds a persistent EMA teacher copy. \our{} needs no extra model copy, since on-policy rollouts reuse the model being trained. (b) Projecting the same accounting to larger models, the absolute gap grows with parameter count.}
    \label{fig:peak_memory}
    \vspace{-6mm}
\end{figure*}

We evaluate \our{} across scientific reasoning and tool-use tasks to determine whether solution-conditioned self-distillation improves learning while eliminating the persistent teacher used by EMA-stabilized on-policy distillation. We compare against verifier-based policy optimization and teacher-based dense supervision under matched wall-clock training budgets.

\vspace{-2mm}
\subsection{Experimental Setting}
\vspace{-2mm}

We evaluate four level-3 reasoning subsets of SciKnowEval~\cite{feng2024sciknoweval}: \textbf{Chemistry}, \textbf{Physics}, \textbf{Biology}, and \textbf{Materials science}, together with \textbf{ToolAlpaca (Tool use)}~\cite{tang2023toolalpaca} for structured tool use. We use Qwen3-8B~\cite{yang2025qwen3} and Olmo3-7B-Instruct~\cite{olmo2025olmo} as initial checkpoints and construct disjoint train--test splits for each task. Performance is measured using avg@16 on the held-out split. We use a single hyperparameter setting across all models and datasets (Appendix~\ref{apen:training_setup}), without dataset specific tuning.


We compare \our{} against GRPO~\cite{shao2024deepseekmath}, on-policy GRPO, and on-policy SDPO~\cite{sdpo}, with baseline results taken from SDPO. All methods use the same initialization and hardware budget. Each run is performed on four NVIDIA GH200 GPUs, and we report the highest accuracy obtained within 1-hour and 5-hour wall-clock training budgets. On-policy GRPO, SDPO, and \our{} perform one gradient update per generation batch, whereas GRPO performs four off-policy mini-batch updates.

\vspace{-2mm}
\subsection{Main Results}
\vspace{-2mm}

We first examine whether the proposed temporal self-distillation strategy translates into consistent improvements across tasks, model backbones, and training budgets. Table~\ref{tab:table1} shows that \our{} achieves the highest Avg@5h for both Qwen3-8B and Olmo3-7B-Instruct, with the best 5-hour performance on every scientific reasoning task. The improvements remain consistent across Chemistry, Physics, Biology, and Materials, suggesting that the gains are not tied to a particular domain or backbone. The margin is smaller on Tool use, where the baselines already reach performance close to ours within the same budget. Even in this setting, \our{} remains competitive at the 5-hour budget, matching the strongest Qwen3-8B result and improving over the baselines on Olmo3-7B-Instruct, although SDPO is stronger at the 1-hour budget on both backbones.

The same behavior becomes clearer when considering optimization over time. As shown in Figure~\ref{fig:convergence}, \our{} obtains a large fraction of its final improvement early in training and approaches its peak substantially faster than the baselines across most tasks. Figure~\ref{fig:time_per_step} further shows that this faster learning is accompanied by a rapid reduction in per-step cost. Privileged supervision and disagreement correction provide stronger local guidance by suppressing overvalued token choices, while temporal consolidation reinforces previously successful solution patterns. As these signals accumulate, the policy increasingly produces shorter responses, which contributes to its lower generation cost. In contrast, GRPO and SDPO tend to generate longer trajectories as training progresses, causing their generation-dominated step time to gradually increase.

\vspace{-3mm}
\subsection{Learning without Forgetting}
\vspace{-2mm}

Improving performance on a target task can degrade behaviors acquired from other domains, making cross-task retention an important measure of stable post-training. Figure~\ref{fig:forgetting} examines this trade-off by comparing new-task gains with performance retained on five prior-capability benchmarks. \our{} improves the trained task while exhibiting consistently higher retention than the baselines. We attribute this to function-space temporal consolidation, which replays fixed historical predictions. This mechanism regularizes updates toward previously useful behavior and reduces catastrophic forgetting during continued adaptation. Detailed cross-task scores are provided in Appendix~\ref{apen:cata_forgetting}.

\vspace{-2mm}
\subsection{Memory and Computational Properties}
\vspace{-2mm}

\begin{wraptable}{r}{0.5\linewidth}
\vspace{-5mm}
\centering
\newcommand{\oc}{\cellcolor{blue!6}}
\caption{Ablation study of \our{} on Olmo3-7B-Instruct across Biology and Tool Use. Shaded rows marked \textit{(ours)} denote the setting used in the full model; unless stated otherwise.}
\vspace{-2mm}
\label{tab:abla}
\Large
\resizebox{\linewidth}{!}{
\begin{tabular}{llccc}
\toprule
\multirow{2}{*}{\textbf{Cases}} & \multirow{2}{*}{\textbf{Setting}}
  & \multicolumn{3}{c}{\textbf{Accuracy (\%)} $\uparrow$} \\
\cmidrule(lr){3-5}
 & & \textbf{Biology} & \textbf{Tool Use} & \textbf{Avg.} \\
\midrule

\multirow{6}{*}{\textbf{Case 1}}
& \oc Full objective \textit{(ours)}                                        & \oc 62.3 & \oc 66.6 & \oc 64.5 \\
& $\mathcal{L}_{\mathrm{distill}}$ only                                     & 54.9 & 57.2 & 56.0 \\
& w/o $\mathcal{L}_{\mathrm{neg}}$                                          & 61.9 & 65.8 & 63.8 \\
& w/o $\mathcal{L}_{\mathrm{temp}}$                                         & 59.9 & 60.3 & 60.1 \\
& w/o $\mathcal{L}_{\mathrm{CE}}$                                           & 59.1 & 64.3 & 61.7 \\
& w/o $\mathcal{L}_{\mathrm{temp}}$ \& $\mathcal{L}_{\mathrm{CE}}$           & 54.4 & 53.8 & 54.1 \\
\midrule

\multirow{4}{*}{\textbf{Case 2}}
& \oc Sibling $\rightarrow$ $\tilde{y}_p$ $\rightarrow$ reference \textit{(ours)} & \oc 62.3 & \oc 66.6 & \oc 64.5 \\
& Reference only                                                            & 41.8 & 39.1 & 40.5 \\
& Verified sibling only                                                     & 57.9 & 60.2 & 59.0 \\
& w/o $\tilde{y}_p$                                                         & 58.8 & 62.6 & 60.7 \\
\midrule

\multirow{6}{*}{\textbf{Case 3}}
& \oc Function-space cache \textit{(ours)}                                  & \oc 62.3 & \oc 66.6 & \oc 64.5 \\
& Parameter snapshot (every $H$ steps)                                      & 59.4 & 58.1 & 58.8 \\
& EMA teacher ($\mu=0.99$)                                                  & 62.3 & 64.2 & 63.2 \\
& EMA teacher ($\mu=0.999$)                                                 & 61.3 & 61.3 & 61.3 \\
& Frozen initial policy                                                     & 56.8 & 56.1 & 56.5 \\
& CE replay of $\tilde{y}_p$                                                & 58.7 & 58.3 & 58.5 \\
\midrule

\multirow{5}{*}{\textbf{Case 4}}
& \oc Quality + coverage retention \textit{(ours)}                          & \oc 62.3 & \oc 66.6 & \oc 64.5 \\
& Reward-only retention                                                     & 59.8 & 64.1 & 61.9 \\
& Coverage-only retention                                                   & 56.3 & 57.8 & 57.0 \\
& Uniform sampling                                                          & 58.9 & 62.4 & 60.6 \\
& Random retention + uniform sampling                                       & 54.8 & 59.4 & 57.1 \\
\midrule

\multirow{3}{*}{\textbf{Case 5}}
& \oc Reliability-adaptive $w_p^{(s)}$ \textit{(ours)}                      & \oc 62.3 & \oc 66.6 & \oc 64.5 \\
& Constant $w_p^{(s)}=1$                                                    & 60.2 & 65.1 & 62.6 \\
& Binary $w_p^{(s)}$                                                        & 60.6 & 65.4 & 63.0 \\
\midrule

\multirow{4}{*}{\textbf{Case 6}}
& \oc Cyclic, adaptation $\rightarrow$ consolidation \textit{(ours)}        & \oc 62.3 & \oc 66.6 & \oc 64.5 \\
& Fixed $\rho_s=0.4$                                                        & 60.6 & 65.1 & 62.8 \\
& Monotonic ramp                                                            & 60.8 & 64.2 & 62.5 \\
& Reverse cyclic, consolidation $\rightarrow$ adaptation                    & 57.6 & 61.7 & 59.7 \\
\midrule
\textbf{Base} & Olmo3-7B-Instruct (zero-shot)                                           & 16.2 & 39.3 & 27.8 \\
\midrule
\rowcolor{blue!15} \textbf{\our{}} & \textbf{Complete Method}               & \textbf{62.3} & \textbf{66.6} & \textbf{64.5} \\
\bottomrule
\end{tabular}}
\vspace{-5mm}
\end{wraptable}

Temporal regularization stabilizes self-distillation, but additional model copies become increasingly costly at scale. \our{} addresses this by storing temporal information as sparse function-space targets rather than as a persistent network. Historical response tokens, selected token probabilities, and pooled tail mass are retained in host memory, while only the current policy remains resident during optimization. Figure~\ref{fig:peak_memory} illustrates the resulting difference. For Olmo3-7B-Instruct, the common training state occupies approximately 56\,GB, whereas GRPO and SDPO reach 70\,GB because of an additional rollout copy and EMA teacher, respectively. Since \our{} reuses the current policy for generation and privileged evaluation, it avoids these persistent model-sized copies, at the cost of additional activations for temporal replay. Under the same accounting, eliminating one \textit{bf16} parameter copy corresponds to roughly 12\,TB of aggregate memory at 6T parameters. It is also worth noting that although the cache grows during training, its peak memory overhead remains negligible at about 5\,MB (Appendix~\ref{apen:training_dynamics}).

\vspace{-2mm}
\subsection{Ablation Study}\label{sec:ablation}
\vspace{-2mm}

We ablate the main components of \our{} on Olmo3-7B-Instruct across Biology and Tool Use to assess their contributions.
As shown in Table~\ref{tab:abla}, we ablate the objective, privileged demonstrations, temporal supervision, memory retention, and adaptive grounding while keeping all other settings fixed.
Together, these components raise average accuracy from 27.8\% for the zero-shot base model to 64.5\%, reflecting their complementary contributions.
Additional analyses of the choices are provided in Appendix~\ref{apen:addi_ablation}.

\textbf{Effective supervision.}
Case~1 first reveals that \our{} benefits from combining complementary forms of supervision rather than relying solely on privileged distillation. Temporal consolidation and reference grounding account for the largest gains, particularly on Tool Use, while disagreement correction provides a smaller but consistent improvement. Importantly, these signals depend on one another: retaining disagreement correction after removing both temporal and reference anchors performs worse than distillation alone, suggesting that aggressive local correction is useful only when the evolving self-teacher is sufficiently stabilized. The source of privileged evidence is equally important. Reference-only supervision leads to a substantial drop, whereas verified siblings and retained trajectories remain closer to the policy's response distribution. Their combination in the full hierarchy therefore provides the strongest current supervision (Case~2).

\textbf{Temporal stability.}
The remaining ablations show that temporal stability depends not simply on retaining past information, but on what is preserved and how it is reused. Sparse function-space targets outperform parameter snapshots, EMA teachers, frozen policies, and trajectory-level CE replay; several of these alternatives even fall below removing temporal consolidation altogether (Case~3). Within the memory itself, retaining trajectories by both quality and coverage performs better than either criterion alone, while uniform or random replay further weakens the historical signal (Case~4). This selective preservation is complemented by reliability-adaptive grounding, which improves over constant or binary reference weighting by adjusting supervision as the policy becomes more reliable (Case~5). Finally, cyclic adaptation followed by consolidation performs best, whereas reversing this order causes a marked degradation, supporting the intended progression from acquiring new privileged information to stabilizing it over time (Case~6).
\vspace{-2mm}
\section{Related Work}
\label{sec:related}
\vspace{-2mm}

Policy-gradient methods \citep{williams1992simple,schulman2017proximal} underpin modern RLVR, with GRPO \citep{shao2024deepseekmath} and its variants \citep{yu2025dapo,liu2025understanding} improving reasoning, coding, tool use, and visual grounding \citep{team2025kimi,lambert2024tulu,el2025competitive,khatri2025art,karim2026focusforcingincontextobject}. However, trajectory-level rewards provide coarse credit when feedback is sparse or uniform across rollouts \citep{kazemnejad2024vineppo,yue2025does}. Prior work addresses this through step-level verifiers \citep{lightman2023let,wang2023math,setlur2024rewarding}, dense reward redistribution \citep{cui2025process,chan2024dense,cao2025scar}, and selective updates at high-entropy decisions \citep{wang2025beyond}. Richer supervision has also emerged through critiques \citep{scheurer2023training,luo2025language,song2026expanding}, self-reflection \citep{shinn2023reflexion,madaan2023self,lee2024reinforcement}, hindsight \citep{andrychowicz2017hindsight,liu2023chain,zhang2023wisdom}, context distillation \citep{snell2022learning,choi2022prompt,kujanpaa2024knowledge}, and verified self-training \citep{zelikman2024star,yang2024self}. On-policy self-distillation brings these ideas together by conditioning the same policy on references \citep{zhao2026self}, demonstrations \citep{shenfeld2026self}, or feedback and successful siblings \citep{sdpo}, and SALD \citep{das2026saldselfreferencedadvantagelearning} extends a similar self-referenced objective, akin to SDPO and SDFT, to text-to-image diffusion models \citep{das2026unityattentionflownetworks,das2026efficienttextguidedconvolutionaladapter,DAS2025113152}; yet the resulting teacher evolves with the student and can suppress useful uncertainty \citep{kim2026why,11491926}. FTSD retains this single-policy setting while introducing temporally fixed historical supervision.

Knowledge distillation transfers teacher distributions to a student \citep{hinton2015distilling,romero2014fitnets}, with sequence-level and on-policy variants extending it to language models \citep{kim2016sequence,gu2023minillm,agarwal2024policy}. Student-generated trajectories reduce train--inference mismatch \citep{ross2011reduction}, while divergence choice governs how teacher behavior is matched \citep{amini2025better}. Evolving targets are commonly stabilized through target networks \citep{atari}, EMA teachers \citep{grill2020bootstrap,tarvainen2017mean}, trust regions \citep{schulman2015trust}, or expert iteration \citep{anthony2017thinking}. In parallel, on-policy adaptation can reduce forgetting \citep{shenfeld2025rl,chen2025retaining}, while continual learning relies on parameter regularization \citep{kirkpatrick2017overcoming}, output distillation \citep{li2016learning}, and experience replay \citep{rolnick2019experience}. Closest to FTSD are function-space approaches that preserve previous outputs \citep{benjamin2019measuring,pan2020continual}, including stored-logit replay \citep{buzzega2020dark}. FTSD adapts this principle to LLM self-distillation through sparse privileged distributions used only during training, rather than memory-augmented inference \citep{behrouz2025titans,cao2025infiniteicl}, learned context representations \citep{eyuboglu2025cartridges}, or persistent model copies common in RL systems \citep{sheng2024hybridflow}.
\vspace{-2mm}
\section{Conclusion}
\label{sec:conclusion}
\vspace{-2mm}

We introduced \our{}, a single-policy self-distillation framework that preserves historical behavior via sparse function-space targets rather than a persistent teacher model. By combining privileged adaptation with disagreement correction, reliability-adaptive grounding, and quality- and coverage-aware temporal memory, \our{} provides stable supervision while retaining verified behaviors across training. Across scientific reasoning and tool-use tasks on two model backbones, \our{} achieves the highest average accuracy under a 5-hour wall-clock budget, converges faster, and better preserves prior capabilities, while reducing estimated peak memory from about 70\,GB to 56\,GB on Olmo3-7B-Instruct. The ablations further show that these gains arise from the interaction between current adaptation, temporal consolidation, and adaptive grounding rather than any single component. Overall, the results suggest that temporal knowledge can be preserved effectively in function space, enabling more memory-efficient and stable self-distillation for continual post-training.

\bibliographystyle{iclr2027_conference}
\bibliography{main}

@inproceedings{agarwal2024policy,
  author    = {Agarwal, Rishabh and Vieillard, Nino and Zhou, Yongchao and Stańczyk, P. and Ramos, Sabela and Geist, Matthieu and Bachem, Olivier},
  title     = {On-Policy Distillation of Language Models: Learning from Self-Generated Mistakes},
  booktitle = {International Conference on Learning Representations},
  year      = {2023},
  eprint    = {2306.13649},
  archivePrefix = {arXiv},
  url       = {https://www.semanticscholar.org/paper/a57ef1f5c3af185af79751855b8033b7fc6d89b3},
}

@inproceedings{amini2025better,
  author    = {Amini, Afra and Vieira, Tim and Cotterell, Ryan},
  title     = {Better Estimation of the Kullback--Leibler Divergence Between Language Models},
  booktitle = {Advances in Neural Information Processing Systems 38},
  year      = {2025},
  doi       = {10.52202/085713-3745},
  url       = {https://doi.org/10.52202/085713-3745},
}

@inproceedings{andrychowicz2017hindsight,
  author    = {Andrychowicz, Marcin and Crow, Dwight and Ray, Alex and Schneider, Jonas and Fong, Rachel and Welinder, Peter and McGrew, Bob and Tobin, Joshua and Abbeel, P. and Zaremba, Wojciech},
  title     = {Hindsight Experience Replay},
  booktitle = {Neural Information Processing Systems},
  year      = {2017},
  eprint    = {1707.01495},
  archivePrefix = {arXiv},
  url       = {https://www.semanticscholar.org/paper/429ed4c9845d0abd1f8204e1d7705919559bc2a2},
}

@inproceedings{anthony2017thinking,
  author    = {Anthony, Thomas W. and Tian, Zheng and Barber, D.},
  title     = {Thinking Fast and Slow with Deep Learning and Tree Search},
  booktitle = {Neural Information Processing Systems},
  year      = {2017},
  eprint    = {1705.08439},
  archivePrefix = {arXiv},
  url       = {https://www.semanticscholar.org/paper/745c614dbd23bc1e3def79f600680b88cee28700},
}

@inproceedings{behrouz2025titans,
  author    = {Behrouz, Ali and Zhong, Peilin and Mirrokni, V.},
  title     = {Titans: Learning to Memorize at Test Time},
  booktitle = {Neural Information Processing Systems},
  year      = {2024},
  eprint    = {2501.00663},
  archivePrefix = {arXiv},
  url       = {https://www.semanticscholar.org/paper/5e7a795d89910634f001cc3a631023f1dd4e2e23},
}

@inproceedings{cao2025infiniteicl,
  author    = {Cao, Bowen and Cai, Deng and Lam, Wai},
  title     = {InfiniteICL: Breaking the Limit of Context Window Size via Long Short-term Memory Transformation},
  booktitle = {Findings of the Association for Computational Linguistics: ACL 2025},
  year      = {2025},
  doi       = {10.18653/v1/2025.findings-acl.595},
  url       = {https://doi.org/10.18653/v1/2025.findings-acl.595},
}

@article{cao2025scar,
  author    = {Cao, Meng and Zhang, Shuyuan and Chang, Xiao-Jun and Precup, D.},
  title     = {SCAR: Shapley Credit Assignment for More Efficient RLHF},
  journal   = {arXiv.org},
  year      = {2025},
  eprint    = {2505.20417},
  archivePrefix = {arXiv},
  url       = {https://www.semanticscholar.org/paper/2109255b50868c453d397bc732fc44d8e83a164f},
}

@inproceedings{chan2024dense,
  author    = {Chan, Alex J. and Sun, Hao and Holt, Samuel and Schaar, M.},
  title     = {Dense Reward for Free in Reinforcement Learning from Human Feedback},
  booktitle = {International Conference on Machine Learning},
  year      = {2024},
  eprint    = {2402.00782},
  archivePrefix = {arXiv},
  url       = {https://www.semanticscholar.org/paper/9732c864d1d4161fcb106f2961d9a80dd4fffc9a},
}

@article{chen2025retaining,
  author    = {Chen, Howard and Razin, Noam and Narasimhan, Karthik R. and Chen, Danqi},
  title     = {Retaining by Doing: The Role of On-Policy Data in Mitigating Forgetting},
  journal   = {arXiv.org},
  year      = {2025},
  eprint    = {2510.18874},
  archivePrefix = {arXiv},
  url       = {https://www.semanticscholar.org/paper/e6f4a5f8a271f3830d0aea0b071986d87ed72999},
}

@article{choi2022prompt,
  author    = {Choi, Eunbi and Jo, Yongrae and Jang, Joel and Seo, Minjoon},
  title     = {Prompt Injection: Parameterization of Fixed Inputs},
  journal   = {arXiv.org},
  year      = {2022},
  eprint    = {2206.11349},
  archivePrefix = {arXiv},
  url       = {http://arxiv.org/pdf/2206.11349},
}

@article{cui2025process,
  author    = {Cui, Ganqu and Yuan, Lifan and Wang, Zefan and Wang, Hanbin and Zhang, Yuchen and Li, Wendi and He, Bingxiang and Fan, Yuchen and Yu, Tianyu and Xu, Qi-Xin and Chen, Weize and Yuan, Jiarui and Chen, Huayu and Zhang, Kaiyan and Lv, Xingtai and Wang, Shuo and Yao, Yuan and Han, Xu and Peng, Hao and Cheng, Yu and Liu, Zhiyuan and Sun, Maosong and Zhou, Bowen and Ding, Ning},
  title     = {Process Reinforcement through Implicit Rewards},
  journal   = {Trans. Mach. Learn. Res.},
  year      = {2025},
  eprint    = {2502.01456},
  archivePrefix = {arXiv},
  url       = {https://www.semanticscholar.org/paper/e23379a9752f57732d311f7a97af2c69af6fae7b},
}

@article{el2025competitive,
  author    = {El-Kishky, Ahmed and Wei, Alexander Z. and Saraiva, Andre and Minaev, Borys and Selsam, Daniel and Dohan, David and Song, Francis and Lightman, H. and Clavera, Ignasi and Pachocki, J. and Tworek, Jerry and Kuhn, Lorenz and Kaiser, Lukasz and Chen, Mark and Schwarzer, Max and Rohaninejad, Mostafa and McAleese, Nat and Contributors, Other and Murk, O. and Garg, Rhythm and Shu, Rui and Sidor, Szymon and Kosaraju, V. and Zhou, Wenda},
  title     = {Competitive Programming with Large Reasoning Models},
  journal   = {arXiv.org},
  year      = {2025},
  eprint    = {2502.06807},
  archivePrefix = {arXiv},
  url       = {https://www.semanticscholar.org/paper/fe2864e9ed7fd1256e52ed17282914d1ddf3fd85},
}

@inproceedings{eyuboglu2025cartridges,
  author    = {Eyuboglu, Sabri and Ehrlich, Ryan and Arora, Simran and Guha, Neel and Zinsley, Dylan and Liu, Emily and Tennien, Will and Rudra, Atri and Zou, James and Mirhoseini, Azalia and Ré, Christopher},
  title     = {Cartridges: Lightweight and general-purpose long context representations via self-study},
  booktitle = {arXiv.org},
  year      = {2025},
  eprint    = {2506.06266},
  archivePrefix = {arXiv},
  url       = {https://www.semanticscholar.org/paper/d1427999c5cd8e9645c81f5dd3c1fb5260514138},
}

@inproceedings{feng2024sciknoweval,
  title={SciKnowEval: Evaluating Multi-level Scientific Knowledge of Large Language Models},
  author={Kehua Feng and Xin-Yi Shen and Weijie Wang and Zhuang Xiang and Yuqi Tang and Qiang Zhang and Keyan Ding},
  year={2024},
  url={https://api.semanticscholar.org/CorpusID:281887235}
}

@inproceedings{grill2020bootstrap,
  author    = {Grill, Jean-Bastien and Strub, Florian and Altch'e, Florent and Tallec, Corentin and Richemond, Pierre H. and Buchatskaya, Elena and Doersch, Carl and Pires, B. '. and Guo, Z. and Azar, M. G. and Piot, Bilal and Kavukcuoglu, K. and Munos, R. and Valko, Michal},
  title     = {Bootstrap Your Own Latent: A New Approach to Self-Supervised Learning},
  booktitle = {Neural Information Processing Systems},
  year      = {2020},
  eprint    = {2006.07733},
  archivePrefix = {arXiv},
  url       = {https://www.semanticscholar.org/paper/38f93092ece8eee9771e61c1edaf11b1293cae1b},
}

@inproceedings{gu2023minillm,
  title={MiniLLM: On-Policy Distillation of Large Language Models},
  author={Yu-Xian Gu and Li Dong and Furu Wei and Minlie Huang},
  year={2023},
  url={https://api.semanticscholar.org/CorpusID:259164722}
}

@article{hinton2015distilling,
  title={Distilling the Knowledge in a Neural Network},
  author={Geoffrey E. Hinton and Oriol Vinyals and Jeffrey Dean},
  journal={ArXiv},
  year={2015},
  volume={abs/1503.02531},
  url={https://api.semanticscholar.org/CorpusID:7200347}
}

@inproceedings{kazemnejad2024vineppo,
  author    = {Kazemnejad, Amirhossein and Aghajohari, Milad and Portelance, Eva and Sordoni, Alessandro and Reddy, Siva and Courville, Aaron and Roux, N. L.},
  title     = {VinePPO: Refining Credit Assignment in RL Training of LLMs},
  booktitle = {International Conference on Machine Learning},
  year      = {2024},
  eprint    = {2410.01679},
  archivePrefix = {arXiv},
  url       = {https://www.semanticscholar.org/paper/ab116ae3e6bc3aedfd1b6ec165e59a8219a5d2f5},
}

@inproceedings{khatri2025art,
  author    = {Khatri, Devvrit and Madaan, Lovish and Tiwari, Rishabh and Bansal, Rachit and Duvvuri, Sai Surya and Zaheer, M. and Dhillon, Inderjit S. and Brandfonbrener, David and Agarwal, Rishabh},
  title     = {The Art of Scaling Reinforcement Learning Compute for LLMs},
  booktitle = {arXiv.org},
  year      = {2025},
  eprint    = {2510.13786},
  archivePrefix = {arXiv},
  url       = {https://www.semanticscholar.org/paper/b1762ae7d8d999213f17856c7f250aa99ab19e5e},
}

@inproceedings{kim2016sequence,
  author    = {Kim, Yoon and Rush, Alexander M.},
  title     = {Sequence-Level Knowledge Distillation},
  booktitle = {Conference on Empirical Methods in Natural Language Processing},
  year      = {2016},
  doi       = {10.18653/v1/D16-1139},
  url       = {https://doi.org/10.18653/v1/D16-1139},
}

@misc{team2025kimi,
  author    = {Team, Kimi and Du, Angang and Gao, Bofei and Xing, Bowei and Jiang, Changjiu and Chen, Cheng and Li, Cheng and Xiao, Chenjun and Du, Chenzhuang and Liao, Chonghua and Tang, Chuning and Wang, Congcong and Zhang, Dehao and Yuan, Enming and Lu, Enzhe and Tang, Feng and Sung, Flood and Wei, Guangda and Lai, Guokun and Guo, Haiqing and Zhu, Han and Ding, Haochen and Hu, Hao-Xing and Yang, Haoming and Zhang, Hao and Yao, Haotian and Zhao, Hao-Dong and Lu, Haoyu and Li, Hao-Ze and Yu, Hao and Gao, Hongcheng and Zheng, Huabin and Yuan, Huan and Chen, Jia and Guo, Jianhang and Su, Jianling and Wang, Jianzhou and Zhao, Jie and Zhang, Jin and Liu, Jingyuan and Yan, Junjie and Wu, Junyan and Shi, Li-Na and Ye, Li-Tao and Yu, Long and Dong, Meng-xiao and Zhang, Neo Y. and Ma, Ning and Pan, Qi and Gong, Qucheng and Liu, Shaowei and Ma, Shen and Wei, Shu-Yan and Cao, S. and Huang, Si-Da and Jiang, Tao and Gao, Wei-Wei and Xiong, Weiming and He, Weiran and Huang, Weixiao and Wu, Wenhao and He, Wen and Wei, Xian-sen and Jia, Xian-Xian and Wu, Xingzhe and Xu, Xinran and Zu, Xinxing and Zhou, Xinyu and Pan, Xue-biao and Charles, Y. and Li, Yang and Hu, Yan-Ling and Liu, Yangyang and Chen, Yanru and Wang, Ye-Jia and Liu, Yi-Bo and Qin, Yidao and Liu, Yifeng and Yang, Yingbo and Bao, Yiping and Du, Yulun and Wu, Yuxin and Wang, Yuzhi and Zhou, Zaida and Wang, Zhaoji and Li, Zhaowei and Zhu, Zhengxin and Zhang, Zheng and Wang, Zhe-Xu and Yang, Zhilin and Huang, Zhiqi and Huang, Zi-Hao and Xu, Ziya and Yang, Zong-Han},
  title     = {Kimi k1.5: Scaling Reinforcement Learning with LLMs},
  journal   = {arXiv.org},
  year      = {2025},
  eprint    = {2501.12599},
  archivePrefix = {arXiv},
  url       = {https://www.semanticscholar.org/paper/668075792a7ab40457d92e09da28d35c879271c3},
}

@misc{kujanpaa2024knowledge,
  author    = {Kujanpää, Kalle and Marttinen, Pekka and Valpola, Harri and Ilin, Alexander},
  title     = {Efficient Knowledge Injection in LLMs via Self-Distillation},
  journal   = {Trans. Mach. Learn. Res.},
  year      = {2024},
  eprint    = {2412.14964},
  archivePrefix = {arXiv},
  url       = {https://www.semanticscholar.org/paper/90d454b3bc86ae8018e79276919e2046e24bc9cd},
}

@inproceedings{lambert2024tulu,
  author    = {Lambert, Nathan and Morrison, Jacob Daniel and Pyatkin, Valentina and Huang, Shengyi and Ivison, Hamish and Brahman, Faeze and Miranda, Lester James Validad and Liu, Alisa and Dziri, Nouha and Lyu, Xinxi and Gu, Yuling and Malik, Saumya and Graf, Victoria and Hwang, Jena D. and Yang, Jiangjiang and Bras, Ronan Le and Tafjord, Oyvind and Wilhelm, Christopher and Soldaini, Luca and Smith, Noah A. and Wang, Yizhong and Dasigi, Pradeep and Hajishirzi, Hanna},
  title     = {TÜLU 3: Pushing Frontiers in Open Language Model Post-Training},
  booktitle = {arXiv.org},
  year      = {2024},
  eprint    = {2411.15124},
  archivePrefix = {arXiv},
  url       = {https://www.semanticscholar.org/paper/6a7c29829227bfd65ae0ffec294a874bb9ea0871},
}

@article{lee2024reinforcement,
  author    = {Lee, Kyungjae and Hwang, Dasol and Park, Sunghyun and Jang, Youngsoo and Lee, Moontae},
  title     = {Reinforcement Learning from Reflective Feedback (RLRF): Aligning and Improving LLMs via Fine-Grained Self-Reflection},
  journal   = {arXiv.org},
  year      = {2024},
  eprint    = {2403.14238},
  archivePrefix = {arXiv},
  url       = {https://www.semanticscholar.org/paper/553c86ba967f55db9caffe08240061b1282da893},
}

@inproceedings{lightman2023let,
  author    = {Lightman, H. and Kosaraju, V. and Burda, Yura and Edwards, Harrison and Baker, Bowen and Lee, Teddy and Leike, Jan and Schulman, John and Sutskever, I. and Cobbe, K.},
  title     = {Let's Verify Step by Step},
  booktitle = {International Conference on Learning Representations},
  year      = {2023},
  eprint    = {2305.20050},
  archivePrefix = {arXiv},
  url       = {http://arxiv.org/pdf/2305.20050},
}

@article{liu2023chain,
  author    = {Liu, Hao and Sferrazza, Carmelo and Abbeel, P.},
  title     = {Chain of Hindsight Aligns Language Models with Feedback},
  journal   = {International Conference on Learning Representations},
  year      = {2023},
  eprint    = {2302.02676},
  archivePrefix = {arXiv},
  url       = {https://arxiv.org/pdf/2302.02676},
}

@inproceedings{liu2025understanding,
  author    = {Liu, Zichen and Chen, Changyu and Li, Wenjun and Qi, Penghui and Pang, Tianyu and Du, Chao and Lee, W. and Lin, Min},
  title     = {Understanding R1-Zero-Like Training: A Critical Perspective},
  booktitle = {arXiv.org},
  year      = {2025},
  eprint    = {2503.20783},
  archivePrefix = {arXiv},
  url       = {https://www.semanticscholar.org/paper/de23d38bc2604dcf334dcc46aff217eb6bcd1fe1},
}

@article{luo2025language,
  author    = {Luo, Renjie and Liu, Zichen and Liu, Xiang-Yan and Du, Chao and Lin, Min and Chen, Wenhu and Lu, Wei and Pang, Tianyu},
  title     = {Language Models Can Learn from Verbal Feedback Without Scalar Rewards},
  journal   = {arXiv.org},
  year      = {2025},
  eprint    = {2509.22638},
  archivePrefix = {arXiv},
  url       = {https://www.semanticscholar.org/paper/df0a07beb75eea9c71ca70344726e5c822974c40},
}

@article{madaan2023self,
  title={Self-refine: Iterative refinement with self-feedback},
  author={Madaan, Aman and Tandon, Niket and Gupta, Prakhar and Hallinan, Skyler and Gao, Luyu and Wiegreffe, Sarah and Alon, Uri and Dziri, Nouha and Prabhumoye, Shrimai and Yang, Yiming and others},
  journal={Advances in neural information processing systems},
  volume={36},
  pages={46534--46594},
  year={2023},
  url={https://proceedings.neurips.cc/paper_files/paper/2023/hash/91edff07232fb1b55a505a9e9f6c0ff3-Abstract-Conference.html}
}

@misc{atari,
  author    = {Mnih, Volodymyr and Kavukcuoglu, K. and Silver, David and Rusu, Andrei A. and Veness, J. and Bellemare, Marc G. and Graves, Alex and Riedmiller, Martin A. and Fidjeland, A. and Ostrovski, Georg and Petersen, S. and Beattie, Charlie and Sadik, Amir and Antonoglou, Ioannis and King, Helen and Kumaran, D. and Wierstra, Daan and Legg, S. and Hassabis, D.},
  title     = {Human-level control through deep reinforcement learning},
  journal   = {Nature},
  year      = {2015},
  doi       = {10.1038/nature14236},
  url       = {https://doi.org/10.1038/nature14236},
}

@inproceedings{olmo2025olmo,
  title={Olmo 3},
  author={Allyson Ettinger and Amanda Bertsch and Bailey Kuehl and David W. Graham and David Heineman and Dirk Groeneveld and Faeze Brahman and Finbarr Timbers and Hamish Ivison and Jacob Daniel Morrison and Jake Poznanski and Kyle Lo and Luca Soldaini and Matt Jordan and Mayee F. Chen and Michael Noukhovitch and Nathan Lambert and Pete Walsh and Pradeep Dasigi and Robert Berry and Saumya Malik and Saurabh Shah and Scott Geng and Shane Arora and Shashank Gupta and Taira Anderson and Teng Xiao and Tyler C. Murray and Tyler Romero and Victoria Graf and Akari Asai and Akshita Bhagia and Alexander Wettig and Alisa Liu and Aman Rangapur and Chloe Anastasiades and Costa Huang and Dustin Schwenk and Harsh M. Trivedi and Ian Magnusson and Jaron Lochner and Jiacheng Liu and Lester James Validad Miranda and Maarten Sap and Malia Morgan and Michaela Schmitz and Michal Guerquin and Michael Wilson and Regan Huff and Ronan Le Bras and Rui Xin and Rulin Shao and Sam Skjonsberg and Shannon Zejiang Shen and Shuyue Stella Li and Tucker Wilde and Valentina Pyatkin and William Merrill and Yapei Chang and Yuling Gu and Zhi-yuan Zeng and Ashish Sabharwal and Luke Zettlemoyer and Pang Wei Koh and Ali Farhadi and Noah A. Smith and Hannaneh Hajishirzi},
  year={2025},
  url={https://api.semanticscholar.org/CorpusID:283908770}
}

@inproceedings{romero2014fitnets,
  author    = {Romero, Adriana and Ballas, Nicolas and Kahou, Samira Ebrahimi and Chassang, Antoine and Gatta, C. and Bengio, Yoshua},
  title     = {FitNets: Hints for Thin Deep Nets},
  booktitle = {International Conference on Learning Representations},
  year      = {2014},
  eprint    = {1412.6550},
  archivePrefix = {arXiv},
  url       = {https://www.semanticscholar.org/paper/8604f376633af8b347e31d84c6150a93b11e34c2},
}

@misc{ross2011reduction,
  author    = {Ross, S. and Gordon, Geoffrey J. and Bagnell, J.},
  title     = {A Reduction of Imitation Learning and Structured Prediction to No-Regret Online Learning},
  journal   = {International Conference on Artificial Intelligence and Statistics},
  year      = {2010},
  eprint    = {1011.0686},
  archivePrefix = {arXiv},
  url       = {https://www.semanticscholar.org/paper/79ab3c49903ec8cb339437ccf5cf998607fc313e},
}

@article{scheurer2023training,
  author    = {Scheurer, J'er'emy and Campos, Jon Ander and Korbak, Tomasz and Chan, Jun Shern and Chen, Angelica and Cho, Kyunghyun and Perez, Ethan},
  title     = {Training Language Models with Language Feedback at Scale},
  journal   = {arXiv.org},
  year      = {2023},
  eprint    = {2303.16755},
  archivePrefix = {arXiv},
  url       = {http://arxiv.org/pdf/2303.16755},
}

@inproceedings{schulman2015trust,
  title={Trust Region Policy Optimization},
  author={John Schulman and Sergey Levine and P. Abbeel and Michael I. Jordan and Philipp Moritz},
  booktitle={International Conference on Machine Learning},
  year={2015},
  url={https://api.semanticscholar.org/CorpusID:16046818}
}

@article{schulman2017proximal,
  title={Proximal Policy Optimization Algorithms},
  author={John Schulman and Filip Wolski and Prafulla Dhariwal and Alec Radford and Oleg Klimov},
  journal={ArXiv},
  year={2017},
  volume={abs/1707.06347},
  url={https://api.semanticscholar.org/CorpusID:28695052}
}

@inproceedings{setlur2024rewarding,
  author    = {Setlur, Amrith Rajagopal and Nagpal, Chirag and Fisch, Adam and Geng, Xinyang and Eisenstein, Jacob and Agarwal, Rishabh and Agarwal, Alekh and Berant, Jonathan and Kumar, Aviral},
  title     = {Rewarding Progress: Scaling Automated Process Verifiers for LLM Reasoning},
  booktitle = {International Conference on Learning Representations},
  year      = {2024},
  eprint    = {2410.08146},
  archivePrefix = {arXiv},
  url       = {https://www.semanticscholar.org/paper/66d9550a51cb493aaeeae72d396b0bdb1ca47fe9},
}

@article{shao2024deepseekmath,
  author    = {Shao, Zhihong and Wang, Peiyi and Zhu, Qihao and Xu, Runxin and Song, Jun-Mei and Zhang, Mingchuan and Li, Y. K. and Wu, Yu and Guo, Daya},
  title     = {DeepSeekMath: Pushing the Limits of Mathematical Reasoning in Open Language Models},
  journal   = {arXiv.org},
  year      = {2024},
  eprint    = {2402.03300},
  archivePrefix = {arXiv},
  url       = {https://www.semanticscholar.org/paper/35b142ea69598e6241f0011312128031df55895c},
}

@article{shenfeld2026self,
  author    = {Shenfeld, Idan and Damani, Mehul and Hübotter, Jonas and Agrawal, Pulkit},
  title     = {Self-Distillation Enables Continual Learning},
  journal   = {arXiv.org},
  year      = {2026},
  eprint    = {2601.19897},
  archivePrefix = {arXiv},
  url       = {https://www.semanticscholar.org/paper/370a293b5dcf9d456ba7f805b5b0a3d8e984ff6e},
}

@inproceedings{shenfeld2025rl,
  author    = {Shenfeld, Idan and Pari, Jyothish and Agrawal, Pulkit},
  title     = {RL's Razor: Why Online Reinforcement Learning Forgets Less},
  booktitle = {arXiv.org},
  year      = {2025},
  eprint    = {2509.04259},
  archivePrefix = {arXiv},
  url       = {https://www.semanticscholar.org/paper/ad74032137b4129b54c5c06be208843d73792e7f},
}

@misc{sheng2024hybridflow,
  author    = {Sheng, Guangming and Zhang, Chi and Ye, Zilingfeng and Wu, Xi-Bin and Zhang, Wang and Zhang, Ru and Peng, Yanghua and Lin, Haibin and Wu, Chuan},
  title     = {HybridFlow: A Flexible and Efficient RLHF Framework},
  journal   = {European Conference on Computer Systems},
  year      = {2024},
  doi       = {10.1145/3689031.3696075},
  url       = {https://doi.org/10.1145/3689031.3696075},
}

@inproceedings{shinn2023reflexion,
  author    = {Shinn, Noah and Cassano, Federico and Labash, Beck and Gopinath, A. and Narasimhan, Karthik and Yao, Shunyu},
  title     = {Reflexion: language agents with verbal reinforcement learning},
  booktitle = {Neural Information Processing Systems},
  year      = {2023},
  doi       = {10.52202/075280-0377},
  url       = {https://doi.org/10.52202/075280-0377},
}

@article{snell2022learning,
  author    = {Snell, Charles Burton and Klein, D. and Zhong, Ruiqi},
  title     = {Learning by Distilling Context},
  journal   = {arXiv.org},
  year      = {2022},
  eprint    = {2209.15189},
  archivePrefix = {arXiv},
  url       = {http://arxiv.org/pdf/2209.15189},
}

@article{song2026expanding,
  author    = {Song, Yu-Da and Chen, Lili and Tajwar, Fahim and Munos, Rémi and Pathak, Deepak and Bagnell, J. and Singh, Aarti and Zanette, Andrea},
  title     = {Expanding the Capabilities of Reinforcement Learning via Text Feedback},
  journal   = {arXiv.org},
  year      = {2026},
  eprint    = {2602.02482},
  archivePrefix = {arXiv},
  url       = {https://www.semanticscholar.org/paper/89e0ab207c62915b5d6d0ae058811bf97fabf171},
}

@article{tang2023toolalpaca,
  author    = {Tang, Qiaoyu and Deng, Zi-Liang and Lin, Hongyu and Han, Xianpei and Liang, Qiao and Cao, Boxi and Sun, Le},
  title     = {ToolAlpaca: Generalized Tool Learning for Language Models with 3000 Simulated Cases},
  journal   = {arXiv.org},
  year      = {2023},
  eprint    = {2306.05301},
  archivePrefix = {arXiv},
  url       = {https://arxiv.org/pdf/2306.05301},
}

@inproceedings{wang2023math,
  author    = {Wang, Peiyi and Li, Lei and Shao, Zhihong and Xu, Runxin and Dai, Damai and Li, Yifei and Chen, Deli and Wu, Yu and Sui, Zhifang},
  title     = {Math-Shepherd: Verify and Reinforce LLMs Step-by-step without Human Annotations},
  booktitle = {Proceedings of the 62nd Annual Meeting of the Association for Computational Linguistics (Volume 1: Long Papers)},
  year      = {2024},
  doi       = {10.18653/v1/2024.acl-long.510},
  url       = {https://doi.org/10.18653/v1/2024.acl-long.510},
}

@inproceedings{wang2025beyond,
  author    = {Wang, Shenzhi and Yu, Le and Gao, Chang and Zheng, Chujie and Liu, Shixuan and Lu, Rui and Dang, Kai and Chen, Xiong-Hui and Yang, Jianxin and Zhang, Zhenru and Liu, Yuqiong and Yang, An and Zhao, Andrew and Yue, Yang and Song, Shiji and Yu, Bowen and Huang, Gao and Lin, Junyang},
  title     = {Beyond the 80/20 Rule: High-Entropy Minority Tokens Drive Effective Reinforcement Learning for LLM Reasoning},
  booktitle = {Advances in Neural Information Processing Systems 38},
  year      = {2025},
  doi       = {10.52202/085713-3850},
  url       = {https://doi.org/10.52202/085713-3850},
}

@misc{williams1992simple,
  author    = {Williams, Ronald J.},
  title     = {Simple Statistical Gradient-Following Algorithms for Connectionist Reinforcement Learning},
  journal   = {Machine-mediated learning},
  year      = {2004},
  doi       = {10.1023/A:1022672621406},
  url       = {https://doi.org/10.1023/A:1022672621406},
}

@misc{yang2025qwen3,
  author    = {Yang, An and Li, An-Feng and Yang, Baosong and Zhang, Beichen and Hui, Binyuan and Zheng, Bo and Yu, Bo-Wen and Gao, Chang and Huang, Chengen and Lv, Chenxu and Zheng, Chujie and Liu, Dayiheng and Zhou, Fan and Huang, Fei and Hu, Feng and Ge, Hao and Wei, Haoran and Lin, Huan and Tang, Jialong and Yang, Jian and Tu, Jianhong and Zhang, Jianwei and Yang, Jianxin and Yang, Jiaxin and Zhou, Jingren and Zhou, Jingren and Lin, Junyan and Dang, K. and Bao, Keqin and Yang, Ke‐Pei and Yu, Le and Deng, Li-Chun and Li, Mei and Xue, Min and Li, Ming-Ze and Zhang, Pei and Wang, Peng and Zhu, Qin and Men, Rui and Gao, Ruize and Liu, Shixuan and Luo, Shuang and Li, Tianhao and Tang, Tian-Yi and Yin, Wenbiao and Ren, Xingzhang and Wang, Xinyu and Zhang, Xinyu and Ren, Xuancheng and Fan, Yang and Su, Yang and Zhang, Yi-Chao and Zhang, Ying-Er and Wan, Yu and Liu, Yuqiong and Wang, Zekun and Cui, Zeyu and Zhang, Zhen-Ru and Zhou, Zhipeng and Qiu, Zihan},
  title     = {Qwen3 Technical Report},
  year      = {2025},
  eprint    = {2505.09388},
  archivePrefix = {arXiv},
  url       = {https://www.semanticscholar.org/paper/d2d84d56f730f81d276a02b48d5d44db5bde0b4a},
}

@inproceedings{yang2024self,
  author    = {Yang, Zhaorui and Pang, Tianyu and Feng, Haozhe and Wang, Han and Chen, Wei and Zhu, Minfeng and Liu, Qian},
  title     = {Self-Distillation Bridges Distribution Gap in Language Model Fine-Tuning},
  booktitle = {Proceedings of the 62nd Annual Meeting of the Association for Computational Linguistics (Volume 1: Long Papers)},
  year      = {2024},
  doi       = {10.18653/v1/2024.acl-long.58},
  url       = {https://doi.org/10.18653/v1/2024.acl-long.58},
}

@misc{yu2025dapo,
      title={DAPO: An Open-Source LLM Reinforcement Learning System at Scale}, 
      author={Qiying Yu and Zheng Zhang and Ruofei Zhu and Yufeng Yuan and Xiaochen Zuo and Yu Yue and Weinan Dai and Tiantian Fan and Gaohong Liu and Lingjun Liu and Xin Liu and Haibin Lin and Zhiqi Lin and Bole Ma and Guangming Sheng and Yuxuan Tong and Chi Zhang and Mofan Zhang and Wang Zhang and Hang Zhu and Jinhua Zhu and Jiaze Chen and Jiangjie Chen and Chengyi Wang and Hongli Yu and Yuxuan Song and Xiangpeng Wei and Hao Zhou and Jingjing Liu and Wei-Ying Ma and Ya-Qin Zhang and Lin Yan and Mu Qiao and Yonghui Wu and Mingxuan Wang},
      year={2025},
      eprint={2503.14476},
      archivePrefix={arXiv},
      primaryClass={cs.LG},
      url={https://arxiv.org/abs/2503.14476}, 
}

@inproceedings{yue2025does,
  author    = {Chen, Zhiqi and Lu, Rui and Zhao, Andrew and Wang, Zhaokai and Yue, Yang and Song, Shiji and Huang, Gao},
  title     = {Does Reinforcement Learning Really Incentivize Reasoning Capacity in LLMs Beyond the Base Model?},
  booktitle = {Advances in Neural Information Processing Systems 38},
  year      = {2025},
  doi       = {10.52202/085713-1933},
  url       = {https://doi.org/10.52202/085713-1933},
}

@inproceedings{zelikman2024star,
  author    = {Zelikman, Eric and Wu, Yuhuai and Mu, Jesse and Goodman, Noah D.},
  title     = {{STaR}: Bootstrapping Reasoning with Reasoning},
  booktitle = {Advances in Neural Information Processing Systems 35},
  year      = {2022},
  doi       = {10.52202/068431-1126},
  url       = {https://doi.org/10.52202/068431-1126},
}

@inproceedings{zhang2023wisdom,
  author    = {Zhang, Tianjun and Liu, Fangchen and Wong, Justin and Abbeel, P. and Gonzalez, Joseph E.},
  title     = {The Wisdom of Hindsight Makes Language Models Better Instruction Followers},
  booktitle = {International Conference on Machine Learning},
  year      = {2023},
  eprint    = {2302.05206},
  archivePrefix = {arXiv},
  url       = {http://arxiv.org/pdf/2302.05206},
}

@article{zhao2026self,
  author    = {Zhao, Siyan and Xie, Zhihui and Liu, Mengchen and Huang, Jing and Pang, Guan and Chen, Feiyu and Grover, Aditya},
  title     = {Self-Distilled Reasoner: On-Policy Self-Distillation for Large Language Models},
  journal   = {arXiv.org},
  year      = {2026},
  eprint    = {2601.18734},
  archivePrefix = {arXiv},
  url       = {https://www.semanticscholar.org/paper/06571d39f14fbd771423bf57c9a237faaf7994bf},
}

@misc{sdpo,
  author    = {Hubotter, Jonas and Lubeck, Frederike and Behric, L. and Baumann, Anton and Bagatella, M. and Marta, Daniel and Hakimi, Ido and Shenfeld, Idan and Buening, Thomas Kleine and Guestrin, Carlos and Krause, Andreas},
  title     = {Reinforcement Learning via Self-Distillation},
  journal   = {arXiv.org},
  year      = {2026},
  eprint    = {2601.20802},
  archivePrefix = {arXiv},
  url       = {https://www.semanticscholar.org/paper/89e83ee3eb95843292b1f75f0e50b89e7667b828},
}

@inproceedings{kim2026why,
  author    = {Kim, Jeonghye and Luo, Xufang and Kim, Minbeom and Lee, Sangmook and Kim, Dohyung and Jeon, Jiwon and Li, Dongsheng and Yang, Yuqing},
  title     = {Why Does Self-Distillation (Sometimes) Degrade the Reasoning Capability of LLMs?},
  booktitle = {arXiv.org},
  year      = {2026},
  eprint    = {2603.24472},
  archivePrefix = {arXiv},
  url       = {https://www.semanticscholar.org/paper/9f5901fa33e9e66791bcf38f83cf9fba5c036187},
}

@inproceedings{buzzega2020dark,
  author    = {Buzzega, Pietro and Boschini, Matteo and Porrello, A. and Abati, Davide and Calderara, S.},
  title     = {Dark Experience for General Continual Learning: a Strong, Simple Baseline},
  booktitle = {Neural Information Processing Systems},
  year      = {2020},
  eprint    = {2004.07211},
  archivePrefix = {arXiv},
  url       = {https://www.semanticscholar.org/paper/e10b7cb072737b1fbdcba7948e573268b3573ae9},
}

@inproceedings{pan2020continual,
  author    = {Pan, Pingbo and Swaroop, S. and Immer, Alexander and Eschenhagen, Runa and Turner, Richard E. and Khan, M. E.},
  title     = {Continual Deep Learning by Functional Regularisation of Memorable Past},
  booktitle = {Neural Information Processing Systems},
  year      = {2020},
  eprint    = {2004.14070},
  archivePrefix = {arXiv},
  url       = {https://www.semanticscholar.org/paper/a61fbcf925485f5b42e0dd185776342862836932},
}

@inproceedings{benjamin2019measuring,
  author    = {Benjamin, Ari S. and Kording, Konrad Paul},
  title     = {Improving generalization by regularizing in $L^2$ function space},
  year      = {2018},
  url       = {https://www.semanticscholar.org/paper/cc6dfb0b9a1c7ff877fba5be80c139b4d4de42e4},
}

@inproceedings{rolnick2019experience,
  author    = {Rolnick, David and Ahuja, Arun and Schwarz, Jonathan and Lillicrap, T. and Wayne, Greg},
  title     = {Experience Replay for Continual Learning},
  booktitle = {Neural Information Processing Systems},
  year      = {2018},
  eprint    = {1811.11682},
  archivePrefix = {arXiv},
  url       = {https://www.semanticscholar.org/paper/d9ff7a9344dd5d6653bd7a02bfd704422bb29951},
}

@article{kirkpatrick2017overcoming,
  author    = {Kirkpatrick, J. and Pascanu, Razvan and Rabinowitz, Neil C. and Veness, J. and Desjardins, Guillaume and Rusu, Andrei A. and Milan, Kieran and Quan, John and Ramalho, Tiago and Grabska-Barwinska, A. and Hassabis, D. and Clopath, C. and Kumaran, D. and Hadsell, R.},
  title     = {Overcoming catastrophic forgetting in neural networks},
  journal   = {Proceedings of the National Academy of Sciences of the United States of America},
  year      = {2016},
  doi       = {10.1073/pnas.1611835114},
  url       = {https://doi.org/10.1073/pnas.1611835114},
}

@ARTICLE{li2016learning,
  author    = {Li, Zhizhong and Hoiem, Derek},
  title     = {Learning without Forgetting},
  journal   = {IEEE Transactions on Pattern Analysis and Machine Intelligence},
  year      = {2016},
  doi       = {10.1007/978-3-319-46493-0_37},
  url       = {https://doi.org/10.1007/978-3-319-46493-0_37},
}

@inproceedings{tarvainen2017mean,
  author    = {Tarvainen, Antti and Valpola, Harri},
  title     = {Mean teachers are better role models: Weight-averaged consistency targets improve semi-supervised deep learning results},
  booktitle = {Neural Information Processing Systems},
  year      = {2017},
  url       = {https://www.semanticscholar.org/paper/7493389667058116dbc7e808987f129325ee60d7},
}

@misc{das2026unityattentionflownetworks,
      title={UNITY: Attention Flow Networks for Adaptive Conditioning in Diffusion}, 
      author={Aryan Das and Koushik Biswas and Moloud Abdar and Vinay Kumar Verma},
      year={2026},
      eprint={2606.20971},
      archivePrefix={arXiv},
      primaryClass={cs.CV},
      url={https://arxiv.org/abs/2606.20971}, 
}

@misc{das2026efficienttextguidedconvolutionaladapter,
      title={Efficient Text-Guided Convolutional Adapter for the Diffusion Model}, 
      author={Aryan Das and Koushik Biswas and Swalpa Kumar Roy and Badri Narayana Patro and Vinay Kumar Verma},
      year={2026},
      eprint={2602.14514},
      archivePrefix={arXiv},
      primaryClass={cs.CV},
      url={https://arxiv.org/abs/2602.14514}, 
}

@article{DAS2025113152,
title = {A generative framework for detection and classification of plant leaf disease using diffusion network},
journal = {Applied Soft Computing},
volume = {177},
pages = {113152},
year = {2025},
issn = {1568-4946},
doi = {https://doi.org/10.1016/j.asoc.2025.113152},
url = {https://www.sciencedirect.com/science/article/pii/S1568494625004636},
author = {Aryan Das and Rajul Mahto and Wentao Wang and Ali Jamali and Pravendra Singh and Swalpa Kumar Roy},
}

@INPROCEEDINGS{11491926,
  author={Das, Aryan and Rachamalla, Tanishq and Biswas, Koushik and Roy, Swalpa Kumar and Kumar Verma, Vinay},
  booktitle={2026 IEEE/CVF Winter Conference on Applications of Computer Vision (WACV)}, 
  title={Uncertainty-Aware Vision-Language Segmentation for Medical Imaging}, 
  year={2026},
  volume={},
  number={},
  pages={8490-8499},
  doi={10.1109/WACV61042.2026.00819}
}

@misc{karim2026focusforcingincontextobject,
      title={FOCUS: Forcing In-Context Object Localization through Visual Support Constraints and Policy Optimization}, 
      author={Mohammed Asad Karim and Vinay Kumar Verma},
      year={2026},
      eprint={2605.31145},
      archivePrefix={arXiv},
      primaryClass={cs.CV},
      url={https://arxiv.org/abs/2605.31145}, 
}

@misc{das2026saldselfreferencedadvantagelearning,
      title={SALD: Self-Referenced Advantage Learning for Diffusion Models}, 
      author={Aryan Das and Surjo Dey and Koushik Biswas and Swalpa Kumar Roy and Moloud Abdar and Arnab Bhattacharya and Vinay Kumar Verma},
      year={2026},
      eprint={2610.01496},
      archivePrefix={arXiv},
      primaryClass={cs.CV},
      url={https://arxiv.org/abs/2610.01496}, 
}

\clearpage
\appendix
\section{Appendix}
\newcommand{\figpanel}[2]{Figure~\hyperref[#1]{\ref*{#1}(#2)}}

\subsection{Complete Training Procedure}
\label{apen:full_algorithm}

Algorithm~\ref{alg:ftsd_full} expands Algorithm~\ref{alg:ftsd} into the complete training loop of \our{}, using the notation of Section~\ref{sec:ftsd}. Each step follows six stages: the policy explores the problem, learns from current privileged evidence, is grounded on the reference solution when it remains unreliable, consolidates fixed historical targets, updates its parameters, and finally refreshes its temporal memory. For readability, $\operatorname{mean}$ denotes the average over valid response positions, matching the normalizers $Z_{\mathrm{cur}}$, $Z_{\mathrm{temp}}$, and $Z_{\mathrm{CE}}$, and $D_{\alpha}$ denotes the same generalized JSD ($\alpha=0.5$) used in $\mathcal{L}_{\mathrm{distill}}$.

Each memory entry holds a retained trajectory $\tilde y_p$, its verifier score $\tilde r_p$, its structural signature $\sigma_p$ (the invoked action sequence for Tool Use, the predicted option for multiple-choice tasks), and its sparse targets $\{\tilde S_{p,t},\tilde q_{p,t}\}_t$. Memory is seeded with reference solutions for the problems in the initial batches, marked with $\tilde r_p=0$ so that they serve only as fallback targets and are not used as verified demonstrations. Verified rollouts are added for problems not yet in memory and replace an existing entry when their retention score, which combines verifier quality with signature coverage using a coverage weight of $0.5$, exceeds that of the stored trajectory.

\begin{algorithm}[!tbh]
\small
\caption{\our{}: Complete Training Procedure}
\label{alg:ftsd_full}
\begin{algorithmic}[1]
\Require Training set $\mathcal{D}=\{(x_p,g_p)\}_{p=1}^{N}$, policy $\pi_{\theta_1}$, verifier $R$, hyperparameters in Table~\ref{tab:train_config}
\State Initialize memory $\mathcal{M}$ with reference solutions for the initial batches and \Call{Store}{} their targets
\For{$s=1,\ldots,T$}
    \State Sample a question batch $\mathcal{B}_s\subset\mathcal{D}$
    \Statex \hspace{\algorithmicindent}\textit{// 1. Explore}
    \State Sample $y_{p,j}\sim\pi_{\theta_s}(\cdot\mid x_p)$ and score $r_{p,j}=R(x_p,y_{p,j},g_p)$ for $j=1,\ldots,G$
    \Statex \hspace{\algorithmicindent}\textit{// 2. Learn from current privileged evidence}
    \State Choose demonstration $d_{p,j}$: verified sibling $\rightarrow$ retained $\tilde y_p$ $\rightarrow$ reference $g_p$
    \State Student $p_{p,j,t}=\pi_{\theta_s}(\cdot\mid x_p,y_{p,j,<t})$, \; teacher $q_{p,j,t}=\operatorname{stopgrad}[\pi_{\theta_s}(\cdot\mid x_p\oplus d_{p,j},y_{p,j,<t})]$
    \State $\mathcal{L}_{\mathrm{distill}}\gets\operatorname{mean}\,\mathrm{JSD}\big(\Phi_{S}(p_{p,j,t}),\Phi_{S}(q_{p,j,t})\big)$, \; $S=\operatorname{TopK}(p_{p,j,t},K_{\mathrm{cur}})$
    \State $\mathcal{L}_{\mathrm{neg}}\gets\operatorname{mean}\,[\log p_{p,j,t}(v)-\log q_{p,j,t}(v)-\gamma]_{+}$ over overvalued tokens $v\in\mathcal{N}_{p,j,t}$
    \State $\mathcal{L}_{\mathrm{cur}}\gets\mathcal{L}_{\mathrm{distill}}+\lambda_{\mathrm{neg}}\mathcal{L}_{\mathrm{neg}}$
    \Statex \hspace{\algorithmicindent}\textit{// 3. Ground on the reference when the policy is unreliable}
    \State $\hat r_p\gets(1-\nu)\,\hat r_p+\nu\cdot\frac{1}{G}\sum_j r_{p,j}$, \; $w_p\gets\operatorname{clip}\big((\tau-\hat r_p)/\tau,0,1\big)$
    \State $\mathcal{L}_{\mathrm{CE}}\gets-\operatorname{mean}\,w_p\log\pi_{\theta_s}(g_{p,t}\mid x_p,g_{p,<t})$
    \Statex \hspace{\algorithmicindent}\textit{// 4. Consolidate fixed historical targets}
    \State Sample $\mathcal{R}_s\subset\mathcal{M}$ evenly across signature groups
    \State $\mathcal{L}_{\mathrm{temp}}\gets\operatorname{mean}\,D_{\alpha}\big(\Phi_{\tilde S_{p,t}}[\pi_{\theta_s}(\cdot\mid x_p,\tilde y_{p,<t})],\tilde q_{p,t}\big)$ for $p\in\mathcal{R}_s$
    \State $\rho_s\gets\rho_{\min}+(\rho_{\max}-\rho_{\min})\frac{(s-1)\bmod C}{\max(1,C-1)}$
    \Statex \hspace{\algorithmicindent}\textit{// 5. Update the policy}
    \State $\mathcal{L}(\theta_s)\gets(1-\rho_s)\mathcal{L}_{\mathrm{cur}}+\rho_s\mathcal{L}_{\mathrm{temp}}+\lambda_{\mathrm{CE}}\mathcal{L}_{\mathrm{CE}}$
    \State $\theta_{s+1}\gets\operatorname{AdamW}\big(\theta_s,\nabla_{\theta}\mathcal{L}(\theta_s)\big)$
    \Statex \hspace{\algorithmicindent}\textit{// 6. Update the temporal memory}
    \For{each $p\in\mathcal{B}_s$ with a verified rollout $y_{p,j^\star}$ ($r_{p,j^\star}\ge\tau_{\mathrm{succ}}$)}
        \If{$p\notin\mathcal{M}$ or $y_{p,j^\star}$ has a higher retention score than $\tilde y_p$}
            \State $(\tilde y_p,\tilde r_p,\sigma_p)\gets(y_{p,j^\star},r_{p,j^\star},\sigma(y_{p,j^\star}))$ and \Call{Store}{$p$}
        \EndIf
    \EndFor
    \If{$s\bmod F=0$}
        \State \Call{Store}{$p$} again for a fraction $\zeta$ of entries older than $H$ steps \Comment{staggered refresh}
    \EndIf
\EndFor
\Statex
\Procedure{Store}{$p$} \Comment{materialize fixed targets with the current policy $\theta_p^{\mathrm{src}}$}
    \State $\tilde d_p\gets\tilde y_p$ if $\tilde r_p\ge\tau_{\mathrm{succ}}$, else $g_p$
    \State $\bar q_{p,t}\gets\operatorname{stopgrad}[\pi_{\theta_p^{\mathrm{src}}}(\cdot\mid x_p\oplus\tilde d_p,\tilde y_{p,<t})]$
    \State $\tilde S_{p,t}\gets\operatorname{TopK}(\bar q_{p,t},K_{\mathrm{temp}})$, \; $\tilde q_{p,t}\gets\Phi_{\tilde S_{p,t}}(\bar q_{p,t})$ \Comment{kept fixed in host memory}
\EndProcedure
\end{algorithmic}
\end{algorithm}

\subsection{Additional Details on Catastrophic Forgetting}
\label{apen:cata_forgetting}

Table~\ref{tab:prior_capabilities} provides a broader view of how task-specific adaptation affects capabilities already present in the base model. Across all five adaptation datasets, standard GRPO produces the largest degradation on the retained benchmarks, indicating substantial drift toward the newly optimized task. On-policy GRPO reduces this degradation, while SDPO further improves retention through dense self-distillation. \our{} consistently provides the strongest overall retention among the adapted models, achieving the highest average score for every training dataset. On the scientific tasks, its average performance remains close to or exceeds the original Olmo3-7B-Instruct checkpoint: Chemistry and Biology reach 60.95 and 60.94 compared with the base score of 61.26, while Physics and Materials improve to 62.61 and 62.32, respectively. These results indicate that preserving historical behavior in function space can limit catastrophic forgetting without preventing beneficial transfer from the newly learned task.

The benchmark-wise results further reveal how this behavior varies across prior capabilities. HumanEval shows the clearest positive transfer under scientific adaptation with \our{}, increasing from 47.56 in the base model to 51.83--58.54 across Chemistry, Physics, Biology, and Materials, with the largest improvement observed after Physics training. MMLU is comparatively stable, remaining close to the original 58.77 and slightly exceeding it after Materials and Tool-use adaptation. In contrast, HellaSwag, Winogrande, and TruthfulQA generally remain below their respective base-model scores, indicating that these capabilities are more susceptible to task-specific adaptation. Tool-use training represents the most challenging setting, where \our{} attains an average retention score of 59.45 compared with 61.26 for the base model. Nevertheless, it still outperforms GRPO (54.79), on-policy GRPO (57.58), and SDPO (58.73) in overall retention. Taken together, these results show that function-space temporal consolidation substantially reduces destructive behavioral drift while still allowing improvements learned during adaptation to transfer to compatible prior capabilities.

\subsection{Additional Ablation Study}
\label{apen:addi_ablation}

\begin{table}[h]
\centering
\caption{
Prior-capability retention on Olmo3-7B-Instruct.
Performance for each method on $5$ prior-capability benchmarks after training on each task dataset.
}
\label{tab:prior_capabilities}

\setlength{\tabcolsep}{4.5pt}
\renewcommand{\arraystretch}{1.12}

\resizebox{\linewidth}{!}{%
\begin{tabular}{cccccccc}
\toprule
\textbf{Method} &
\textbf{Dataset} &
\textbf{HumanEval} &
\textbf{MMLU} &
\textbf{HellaSwag} &
\textbf{Winogrande} &
\textbf{TruthfulQA} &
\textbf{Avg.} \\
\midrule

\textbf{Olmo3-7B-Instruct}
& --
& 47.56
& 58.77
& \textbf{76.04}
& \textbf{66.14}
& \textbf{57.79}
& 61.26 \\

\midrule

\multirow{5}{*}{GRPO}
& Chemistry & 43.90 & 54.18 & 71.42 & 62.51 & 53.37 & 57.08 \\
& Physics   & 44.51 & 53.95 & 70.88 & 63.14 & 53.92 & 57.28 \\
& Biology   & 42.68 & 54.58 & 71.95 & 63.69 & 53.14 & 57.21 \\
& Materials & 43.29 & 53.82 & 70.61 & 62.75 & 53.85 & 56.86 \\
& Tool use  & 38.41 & 52.94 & 69.95 & 60.77 & 51.88 & 54.79 \\

\midrule

\multirow{5}{*}{GRPO (on-policy)}
& Chemistry & 45.73 & 56.28 & 74.31 & 64.25 & 56.08 & 59.33 \\
& Physics   & 44.51 & 56.75 & 74.62 & 64.33 & 56.41 & 59.32 \\
& Biology   & 45.12 & 55.98 & 73.85 & 63.77 & 55.72 & 58.89 \\
& Materials & 46.34 & 56.52 & 74.24 & 64.17 & 56.33 & 59.52 \\
& Tool use  & 40.85 & 55.37 & 73.42 & 63.30 & 54.95 & 57.58 \\

\midrule

\multirow{5}{*}{SDPO (on-policy)}
& Chemistry & 46.95 & 57.48 & 73.82 & 65.19 & 56.42 & 59.97 \\
& Physics   & 45.73 & 57.92 & 74.35 & 64.88 & 56.85 & 59.95 \\
& Biology   & 46.34 & 57.15 & 73.58 & 64.64 & 56.51 & 59.64 \\
& Materials & 46.95 & 57.68 & 74.12 & 65.27 & 56.92 & 60.19 \\
& Tool use  & 43.29 & 56.82 & 73.65 & 64.01 & 55.88 & 58.73 \\

\midrule

\multirow{5}{*}{\textbf{\our{} (on-policy)}}
& Chemistry
& 51.83 & 58.23 & 74.59 & 64.80 & 55.32 & 60.95 \\

& Physics
& \textbf{58.54} & 57.93 & 74.47 & 64.80 & 57.31 & \textbf{62.61} \\

& Biology
& 51.83 & 56.90 & 74.79 & 64.96 & 56.22 & 60.94 \\

& Materials
& 56.10 & \textbf{58.97} & 74.64 & 64.80 & 57.07 & 62.32 \\

& Tool use
& 42.68 & 58.87 & 75.63 & 64.96 & 55.11 & 59.45 \\

\bottomrule
\end{tabular}%
}

\end{table}

\begin{table}[h]
\centering
\newcommand{\oc}{\cellcolor{blue!6}}
\caption{Additional ablations of \our{} on Olmo3-7B-Instruct across Biology and Tool Use. Shaded rows marked \textit{(ours)} denote the setting used in the full model; unless stated otherwise, all other settings follow the full model.}
\label{tab:abla_app}
\resizebox{0.7\linewidth}{!}{
\begin{tabular}{llccc}
\toprule
\multirow{2}{*}{\textbf{Cases}} & \multirow{2}{*}{\textbf{Setting}}
  & \multicolumn{3}{c}{\textbf{Accuracy (\%)} $\uparrow$} \\
\cmidrule(lr){3-5}
 & & \textbf{Biology} & \textbf{Tool Use} & \textbf{Avg.} \\
\midrule
\multirow{3}{*}{\textbf{Case 7}}
& $G=4$                                                                     & 59.6 & 63.9 & 61.8 \\
& \oc $G=8$ \textit{(ours)}                                                 & \oc 62.3 & \oc 66.6 & \oc 64.5 \\
& $G=16$                                                                    & 61.4 & 66.7 & 64.0 \\
\midrule
\multirow{5}{*}{\textbf{Case 8}}
& Forward KL                                                                & 59.4 & 64.2 & 61.8 \\
& JSD, $\alpha=0.25$                                                        & 60.8 & 65.3 & 63.0 \\
& \oc JSD, $\alpha=0.5$ \textit{(ours)}                                     & \oc 62.3 & \oc 66.6 & \oc 64.5 \\
& JSD, $\alpha=0.75$                                                        & 61.1 & 65.9 & 63.5 \\
& Reverse KL                                                                & 60.2 & 62.7 & 61.5 \\
\midrule
\multirow{8}{*}{\textbf{Case 9}}
& $\gamma=0.0$                                                              & 60.9 & 64.1 & 62.5 \\
& $\gamma=0.3$                                                              & 61.6 & 66.2 & 63.9 \\
& \oc $\gamma=0.5$ \textit{(ours)}                                          & \oc 62.3 & \oc 66.6 & \oc 64.5 \\
& $\gamma=1.0$                                                              & 61.3 & 64.8 & 63.0 \\
\cmidrule(lr){2-5}
& $\lambda_{\mathrm{neg}}=0.05$                                             & 61.7 & 66.3 & 64.0 \\
& \oc $\lambda_{\mathrm{neg}}=0.1$ \textit{(ours)}                          & \oc 62.3 & \oc 66.6 & \oc 64.5 \\
& $\lambda_{\mathrm{neg}}=0.5$                                              & 62.3 & 65.3 & 63.8 \\
& $\lambda_{\mathrm{neg}}=1.0$                                              & 59.4 & 62.2 & 60.8 \\
\midrule
\multirow{9}{*}{\textbf{Case 10}}
& Immediate refresh ($H=1$)                                                 & 56.6 & 57.3 & 57.0 \\
& $H=2$                                                                     & 60.3 & 63.9 & 62.1 \\
& \oc $H=4$ \textit{(ours)}                                                 & \oc 62.3 & \oc 66.6 & \oc 64.5 \\
& $H=8$                                                                     & 58.8 & 63.3 & 61.0 \\
& No refresh ($H=\infty$)                                                   & 58.4 & 61.8 & 60.1 \\
\cmidrule(lr){2-5}
& $\zeta=0.1$                                                               & 60.3 & 64.6 & 62.4 \\
& $\zeta=0.25$                                                              & 60.8 & 63.8 & 62.3 \\
& \oc $\zeta=0.5$ \textit{(ours)}                                           & \oc 62.3 & \oc 66.6 & \oc 64.5 \\
& Synchronous refresh ($\zeta=1$)                                           & 60.2 & 63.2 & 61.7 \\
\midrule
\multirow{12}{*}{\textbf{Case 11}}
& $\nu=0.1$                                                                 & 60.4 & 64.6 & 62.5 \\
& \oc $\nu=0.3$ \textit{(ours)}                                             & \oc 62.3 & \oc 66.6 & \oc 64.5 \\
& $\nu=0.5$                                                                 & 61.3 & 64.6 & 62.9 \\
& No smoothing ($\nu=1$)                                                    & 58.8 & 63.4 & 61.1 \\
\cmidrule(lr){2-5}
& $\tau=0.25$                                                               & 61.3 & 65.9 & 63.6 \\
& \oc $\tau=0.6$ \textit{(ours)}                                            & \oc 62.3 & \oc 66.6 & \oc 64.5 \\
& $\tau=0.75$                                                               & 61.3 & 64.8 & 63.0 \\
& $\tau=1.0$                                                                & 59.3 & 63.4 & 61.3 \\
\cmidrule(lr){2-5}
& $\lambda_{\mathrm{CE}}=0.05$                                              & 60.2 & 64.1 & 62.1 \\
& \oc $\lambda_{\mathrm{CE}}=0.1$ \textit{(ours)}                           & \oc 62.3 & \oc 66.6 & \oc 64.5 \\
& $\lambda_{\mathrm{CE}}=0.5$                                               & 58.4 & 60.4 & 59.4 \\
& $\lambda_{\mathrm{CE}}=1.0$                                               & 54.8 & 55.8 & 55.3 \\
\midrule
\multirow{7}{*}{\textbf{Case 12}}
& $C=2$                                                                     & 60.2 & 63.7 & 62.0 \\
& \oc $C=5$ \textit{(ours)}                                                 & \oc 62.3 & \oc 66.6 & \oc 64.5 \\
& $C=10$                                                                    & 61.9 & 65.9 & 63.9 \\
\cmidrule(lr){2-5}
& $(\rho_{\min},\rho_{\max})=(0.1,0.3)$                                     & 60.1 & 64.2 & 62.2 \\
& \oc $(\rho_{\min},\rho_{\max})=(0.3,0.5)$ \textit{(ours)}                 & \oc 62.3 & \oc 66.6 & \oc 64.5 \\
& $(\rho_{\min},\rho_{\max})=(0.5,0.7)$                                     & 60.8 & 63.4 & 62.1 \\
& $(\rho_{\min},\rho_{\max})=(0.1,0.7)$                                     & 61.8 & 65.4 & 63.6 \\
\bottomrule
\end{tabular}}
\end{table}

\begin{table}[t]
\centering
\caption{Training configuration of \our{}.}
\label{tab:train_config}
\small
\resizebox{0.8\linewidth}{!}{
\begin{tabular}{ll}
\toprule
\textbf{Parameters} & \textbf{Value} \\
\midrule
\multicolumn{2}{l}{\textbf{General}} \\
Model            & Qwen/Qwen3-8B \\
                 & allenai/Olmo3-7B-Instruct \\
Thinking         & False \\
\midrule
\multicolumn{2}{l}{\textbf{Data}} \\
Max. prompt length                   & 2048 \\
Max. reprompt (teacher) length       & 10240 \\
Max. response length                 & 8192 \\
\midrule
\multicolumn{2}{l}{\textbf{Batching}} \\
Question batch size   & 32 \\
Mini batch size       & 32 \\
Number of rollouts    & 8 \\
\midrule
\multicolumn{2}{l}{\textbf{Rollout}} \\
Inference engine      & HF Transformers (resident policy) \\
Temperature           & 1.0 \\
Top-$p$               & 1.0 \\
\midrule
\multicolumn{2}{l}{\textbf{Validation}} \\
Inference engine      & vLLM \\
Number of rollouts    & 16 \\
Temperature           & 0.6 \\
Top-$p$               & 0.95 \\
\midrule
\multicolumn{2}{l}{\textbf{Distillation loss}} \\
Top-$K$ distillation ($K_{\mathrm{cur}}$)   & 100 \\
Distillation divergence                    & Jensen--Shannon ($\alpha=0.5$) \\
\midrule
\multicolumn{2}{l}{\textbf{Training}} \\
Optimizer            & AdamW ($\beta_1=0.9$, $\beta_2=0.999$) \\
Learning rate        & $1\times10^{-5}$ \\
Warmup steps         & 10 \\
Weight decay         & 0.01 \\
Gradient clip norm   & 1.0 \\
Precision            & bf16 \\
\midrule
\multicolumn{2}{l}{\textbf{\our{}-specific}} \\
Temporal cache top-$K$ ($K_{\mathrm{temp}}$)          & 32 \\
Success threshold $\tau_{\mathrm{succ}}$              & 0.5 \\
Refresh delay $H$                                     & 4 \\
Staggered refresh period $F$ / fraction $\zeta$       & 10 / 0.5 \\
Coverage weight in memory retention                   & 0.5 \\
Replayed trajectories per step                        & 1$\times$ local batch \\
Disagreement margin $\gamma$ / weight $\lambda_{\mathrm{neg}}$ & 0.5 / 0.1 \\
Grounding weight $\lambda_{\mathrm{CE}}$              & 0.1 \\
Target reliability $\tau$ / smoothing $\nu$           & 0.6 / 0.3 \\
Schedule range $(\rho_{\min}, \rho_{\max})$ / cycle $C$ & (0.3, 0.5) / 5 \\
\bottomrule
\end{tabular}
}
\end{table}

\begin{figure*}[t]
  \centering
  \includegraphics[width=\textwidth]{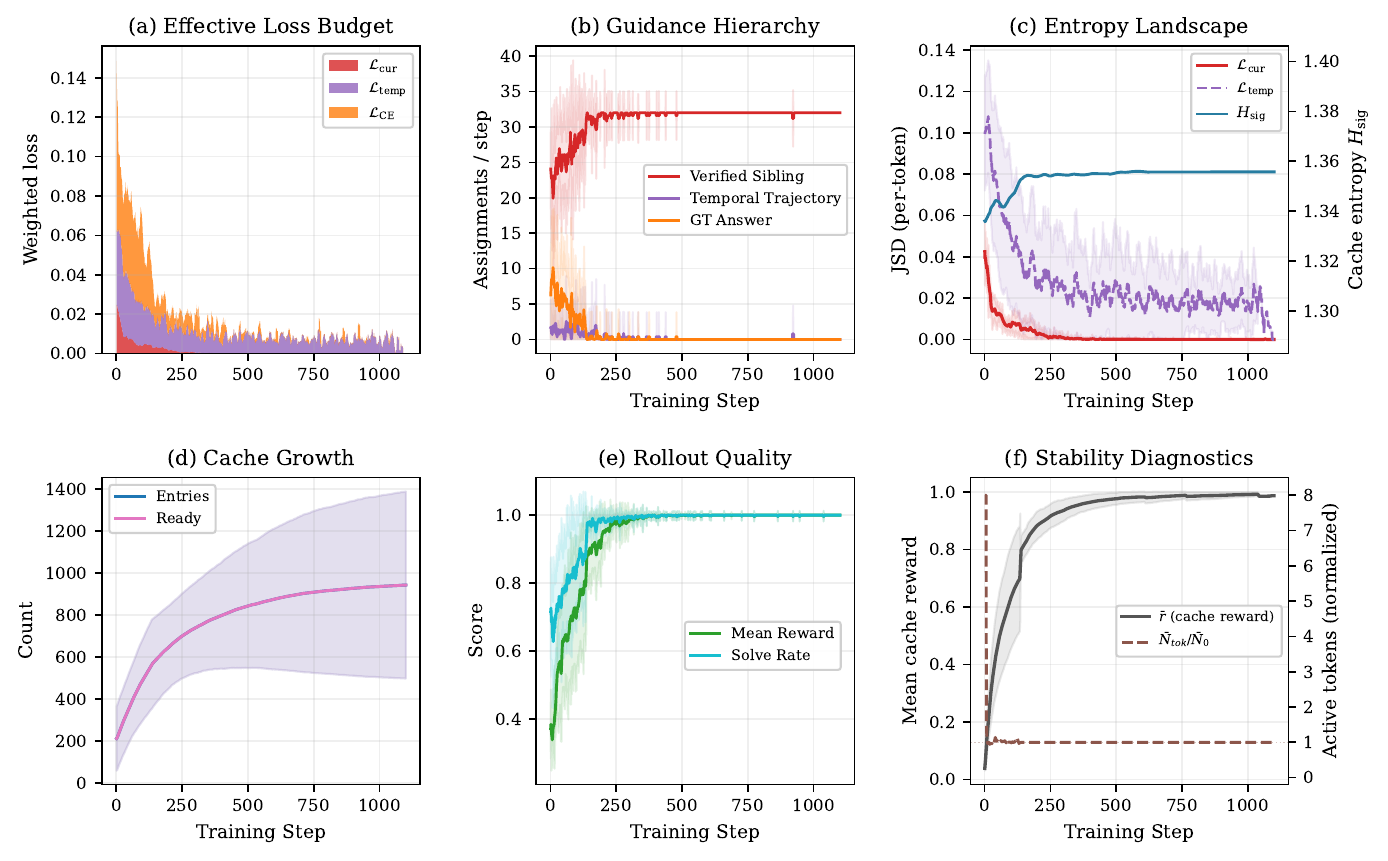}
  \caption{
      Training dynamics of \our{} averaged across five tasks on Olmo3-7B-Instruct.
      Metrics are reported as mean $\pm$ 1\,$\sigma$ over Biology, Chemistry, Materials, Physics, and Tool Use. When a task's training completes before the longest run, its final values are held constant so that all five tasks contribute to the average at every step.
      (a) Weighted contributions of the current, temporal, and reference-grounding objectives under the cyclic $\rho$ schedule.
      (b) Number of guidance-tier assignments per training step.
      (c) Per-token JSD divergence for the current and temporal components, with cache signature entropy $H_{\mathrm{sig}}$ (right axis; science tasks only).
      (d) Growth of the temporal cache, showing total and ready entries.
      (e) Mean rollout reward and per-problem solve rate.
      (f) Mean cache reward and active-token length normalized to its steady-state baseline.
  }
  \label{fig:training_dynamics}
\end{figure*}

\begin{figure*}[t]
  \centering
  \includegraphics[width=0.9\textwidth]{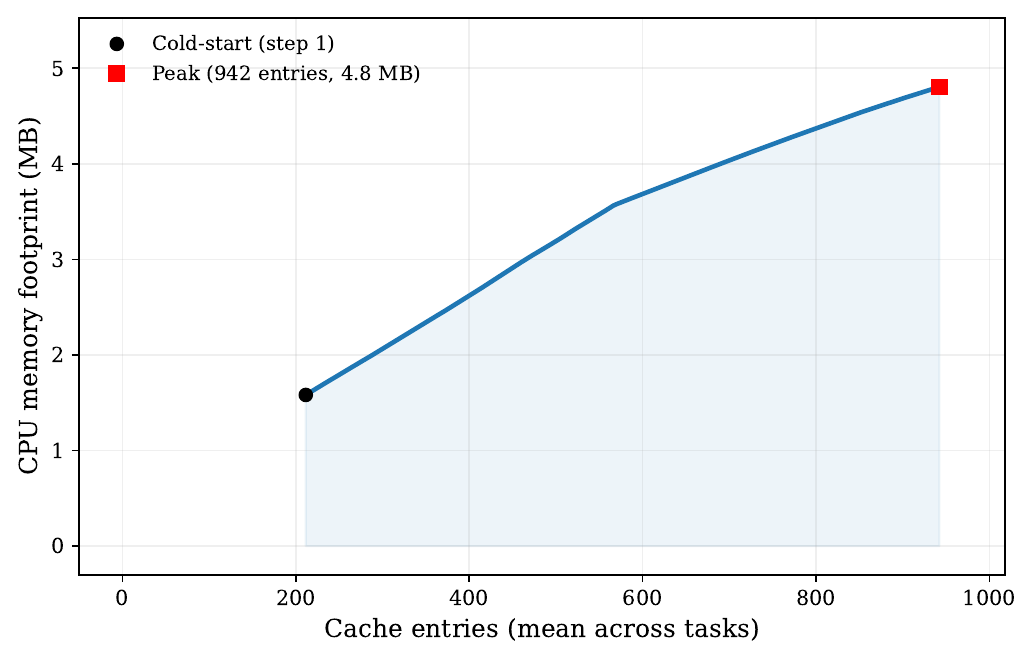}
  \caption{
  Memory footprint of the temporal cache during training.
  CPU memory of ready cache entries plotted against cache size, averaged across the same five tasks as in \figpanel{fig:training_dynamics}{d}. The black marker denotes the seeded cache at step~1, and the red marker denotes the largest observed cache. Each individual task cache is non-decreasing; completed runs are held at their final value so that the cross-task average grows monotonically.
  }
  \label{fig:cache_memory_footprint}
\end{figure*}

Table~\ref{tab:abla_app} continues the ablation study of Section~3.5 with sensitivity analyses over the rollout group size, matching objective, disagreement correction, target refresh, reference grounding, and schedule parameters. As in Table~\ref{tab:abla}, each case modifies a single factor while keeping all other settings identical to the full model.

\textbf{Rollout size and matching objective.}
A group size of $G=8$ provides enough rollouts to find verified siblings for most problems, while $G=16$ yields no further gain under the fixed wall-clock budget (Case 7). The matching objective consistently favors intermediate choices: symmetric JSD outperforms both forward and reverse KL (Case 8), and a moderate correction margin and weight help, whereas aggressive correction ($\lambda_{\mathrm{neg}}=1.0$) falls below removing $\mathcal{L}_{\mathrm{neg}}$ entirely (Case 9).

\textbf{Target refresh.}
Refresh exhibits a trade-off between freshness and temporal separation. Refreshing targets too early ($H=1$) weakens temporal separation, absent refresh leaves targets stale, and a delayed, staggered update ($H=4$, $\zeta=0.5$) performs best (Case 10).

\textbf{Grounding and schedule parameters.}
Smoothing the reliability estimate is important, as reacting to single-step outcomes ($\nu=1$) degrades performance, and an overly strict target reliability ($\tau=1.0$) keeps grounding active on problems the policy already solves. Strong grounding sharply degrades performance by overriding useful self-generated behavior, with $\lambda_{\mathrm{CE}}=1.0$ falling close to the $\mathcal{L}_{\mathrm{distill}}$-only variant (Case 11). For the schedule, a moderate cycle length and consolidation range work best, while the wide range $(0.1, 0.7)$ remains competitive (Case 12).

\subsection{Training Configuration}
\label{apen:training_setup}

Table~\ref{tab:train_config} lists the training configuration of \our{}. We follow the SDPO setting without environment feedback~\citep{sdpo}, from which our baseline results are taken, for the backbones, data limits, batching, rollout sampling, validation protocol, distillation objective, and optimizer, so that the comparison isolates the effect of the proposed objective. Because \our{} performs rollouts and privileged evaluation with the single resident policy, it requires neither an EMA teacher nor rollout importance-sampling correction. Hyperparameters specific to \our{} are listed in the lower block and follow the full-model settings of Table~\ref{tab:abla}.

\subsection{Training Dynamics}
\label{apen:training_dynamics}

Final performance alone does not reveal how the components of \our{} interact during optimization. We therefore examine the training dynamics in Figure~\ref{fig:training_dynamics}, following how supervision evolves as the policy moves from unreliable early rollouts toward stable self-generated guidance. The analysis tracks how the strength and source of supervision change over time, how temporal disagreement is preserved, how the cache matures, and how these mechanisms translate into improved rollout quality and more stable generations. Figure~\ref{fig:cache_memory_footprint} then complements this view by showing that the historical memory supporting temporal consolidation remains negligible in cost even as the cache grows.

\figpanel{fig:training_dynamics}{a} first shows how the effective loss budget changes over this transition. Reference grounding contributes strongly at the beginning, when rollout quality is still low and the adaptive weight $w_p^{(s)}$ therefore places greater emphasis on $\mathcal{L}_{\mathrm{CE}}$. Its contribution then falls rapidly as the policy begins to produce reliable evidence on its own. The current privileged objective follows a similar decline as the student aligns with its contemporaneous self-teacher. Temporal consolidation, in contrast, remains active for longer because cached targets were produced by earlier policy states and therefore continue to disagree with the evolving student. The smaller recurrent fluctuations later in training reflect the cyclic $\rho_s$ schedule, which repeatedly rebalances adaptation and consolidation rather than fixing their relative importance throughout optimization.

The source of this privileged guidance evolves accordingly in \figpanel{fig:training_dynamics}{b}. Verified siblings already account for a substantial portion of guidance early in training and soon become dominant as successful rollouts become more frequent. Reference answers serve mainly as a fallback during the least reliable stage, but their use quickly declines once the policy can generate sufficiently good evidence itself. Retained temporal trajectories occupy the intermediate tier of the hierarchy: they are selected when the current rollout group lacks a verified sibling but useful historical behavior is available in memory. Their contribution therefore remains smaller and gradually becomes rare. The occasional later spikes in the fallback tiers correspond to isolated batches in which verified siblings are unavailable, rather than a reversal of the broader movement toward self-supported supervision.

\figpanel{fig:training_dynamics}{c} clarifies why temporal supervision remains useful after current adaptation has mostly converged. The current JSD falls rapidly toward zero as the unassisted and privileged predictions become aligned at the same policy state. Temporal JSD starts substantially higher and decays more slowly, preserving a nontrivial gap long after the current divergence becomes small. This sustained mismatch is precisely what temporal consolidation is meant to preserve: cached distributions remain fixed while the policy continues to evolve, so the student continues to receive a historical correction signal until those targets are refreshed. At the same time, the cache signature entropy $H_{\mathrm{sig}}$ rises during early memory formation and then stabilizes. Since the signature on these tasks is the predicted option, the plateau near $\ln 4$ indicates that the memory remains balanced across answer options rather than collapsing toward a single one.

The memory underlying this process is shown directly in \figpanel{fig:training_dynamics}{d}. The cache does not start empty: before the first update it is seeded with reference trajectories and sparse teacher targets, so temporal consolidation is active from the beginning of training. From this initial pool, the cache grows steadily within each task as new verified behaviors are admitted or existing entries are replaced, while the number of ready entries closely tracks the total count. This close coupling indicates that most retained experiences quickly become usable temporal targets. The gradual flattening of the averaged curve as tasks complete simply reflects that finished runs hold their final cache size while the remaining tasks continue to grow. The narrowing variance band likewise reflects the convergence of cache sizes across tasks as each run reaches its dataset limit.

Figure~\ref{fig:cache_memory_footprint} translates this expansion into actual memory cost. As the number of retained entries increases, the footprint grows gradually from the seeded cache at step~1 to 942 mean entries and $4.8$\,MB across five tasks. The growth remains small because each entry stores only sparse function-space information, including trajectory tokens, a top-$K$ teacher distribution, and pooled tail mass, rather than any model-sized state. Each task's cache size is non-decreasing; completed runs are held at their final value so the averaged curve reflects cumulative growth across all five tasks. Even at its peak, the cache occupies only about $5$\,MB of CPU memory, confirming that the temporal memory remains negligible relative to the resident model state.

The effect of these mechanisms on policy quality becomes visible in \figpanel{fig:training_dynamics}{e}. Both mean rollout reward and solve rate improve sharply during the early phase, but the solve rate rises faster and saturates sooner. This separation follows from how the two quantities are defined: a problem is counted as solved once one successful rollout appears within its group, whereas mean reward averages performance across all $G$ sampled trajectories. Early in training, the policy can therefore discover one correct trajectory while still producing several unsuccessful alternatives, yielding a high solve rate but a lower average reward. As training progresses, successful behavior becomes more consistent across the rollout group, narrowing this gap and reducing the variance across tasks. The resulting plateau indicates that rapid early improvement gives way to a stable regime rather than continued oscillation.

Finally, \figpanel{fig:training_dynamics}{f} shows how the quality of retained behavior and generation structure evolve together. Mean cache reward rises rapidly as stronger verified trajectories replace weaker initial memories and then approaches a stable high-quality regime. Active-token length follows a different pattern: it exhibits a brief spike at the very beginning, when the unadapted policy can generate substantially longer trajectories, but then quickly contracts toward its steady-state level and remains stable thereafter. This sharp early correction suggests that the policy is learning to produce more task-sufficient responses rather than simply generating longer ones. Together with the rise in cache reward, this indicates that improved solution quality emerges alongside increasingly compact and stable generations, rather than being driven by uncontrolled expansion of trajectory length.

\end{document}